\documentclass[letterpaper, 10 pt, conference]{ieeeconf}  

\IEEEoverridecommandlockouts                              

\usepackage{times,url,amsmath,graphicx,algorithm,algorithmic,hyperref}
\usepackage{color}

\long\def\ignorethis#1{}

\newcommand{\ie}{i.e.\ }

\newcommand{\tr}{^\mathrm{T}}

\newcommand{\gauss}{\mathcal{N}}

\newcommand{\trajdist}{p}
\newcommand{\policy}{\pi}

\newcommand{\params}{\theta}

\newcommand{\cost}{\ell}
\newcommand{\state}{\mathbf{x}}
\newcommand{\obs}{\mathbf{o}}
\newcommand{\action}{\mathbf{u}}

\newcommand{\traj}{\tau}

\newcommand{\ucovart}{\mathbf{C}_t}

\newcommand{\kl}{D_\text{KL}}

\newcommand{\fct}{f_{c t}}
\newcommand{\fxt}{f_{\state t}}
\newcommand{\fut}{f_{\action t}}

\newcommand{\kpol}{\mathbf{k}}
\newcommand{\Kpol}{\mathbf{K}}

\newcommand{\noise}{\mathbf{F}}

\newcommand{\hidact}{\mathbf{a}}
\newcommand{\hidstate}{\mathbf{h}}

\newcommand{\rnnout}{\phi}
\newcommand{\rnndyn}{\psi}

\newcommand{\vat}{\hidact_t}
\newcommand{\vt}{\hidstate_t}
\newcommand{\st}{\state_t}
\newcommand{\tstate}{\tilde{\state}}
\newcommand{\taction}{\tilde{\action}}
\newcommand{\tst}{\tilde{\state}_t}
\newcommand{\tat}{\tilde{\action}_t}
\newcommand{\tot}{\tilde{\obs}_t}
\newcommand{\ot}{\obs_t}
\newcommand{\at}{\action_t}

\newcommand{\lgmut}{\lambda_{\mu t}}
\newcommand{\admmrho}{\nu}

\title{\LARGE \bf
Learning Deep Neural Network Policies with Continuous Memory States
}

\author{
Marvin Zhang, Zoe McCarthy, Chelsea Finn, Sergey Levine, Pieter Abbeel%
\thanks{Department of Electrical Engineering and Computer Science, University of
    California, Berkeley, Berkeley, CA 94709}%
}

\begin{document}

\maketitle
\thispagestyle{empty}
\pagestyle{empty}

\begin{abstract}

Policy learning for partially observed control tasks requires policies that can
remember salient information from past observations. In this paper, we present a
method for learning policies with internal memory for high-dimensional,
continuous systems, such as robotic manipulators. Our approach consists of
augmenting the state and action space of the system with continuous-valued
memory states that the policy can read from and write to. Learning
general-purpose policies with this type of memory representation directly is
difficult, because the policy must automatically figure out the most salient
information to memorize at each time step. We show that, by decomposing this
policy search problem into a trajectory optimization phase and a supervised
learning phase through a method called guided policy search, we can acquire
policies with effective memorization and recall strategies. Intuitively, the
trajectory optimization phase chooses the values of the memory states that will
make it easier for the policy to produce the right action in future states,
while the supervised learning phase encourages the policy to use memorization
actions to produce those memory states. We evaluate our method on tasks
involving continuous control in manipulation and navigation settings, and show
that our method can learn complex policies that successfully complete a range of
tasks that require memory.

\end{abstract}

\section{Introduction}

Reinforcement learning (RL) and optimal control methods have the potential to
allow robots to autonomously discover complex behaviors. However, robotic
control problems are often continuous, high dimensional, and partially observed.
The partial observability in particular presents a major challenge. Partial
observability has been tackled in the context of POMDPs by using a variety of
model-based approximations \cite{shani2013survey}. However, despite recent
progress
\cite{meuleau1999learning,peshkin2001learning,toussaint2008hierarchical,deisenroth2011pilco,wsbr-e2c-15},
learning the state representation, the dynamics and the observation model
together remains challenging. Model-free policy search algorithms have been
successfully used to sidestep the need for learning dynamics and observation
models, by optimizing policies directly through system interaction
\cite{dnp-spsr-13}. However, partially observed domains, where reactive policies
are insufficient, necessitate the use of internal memory. Finite state
controllers have previously been applied to smaller RL tasks where value
function approximation is practical \cite{meuleau1999learning}, and policy
gradient methods have been extended to recurrent neural networks (RNNs)
\cite{wierstra2007solving}. However, effective training of complex,
high-dimensional, general-purpose policies with internal memory still presents a
tremendous challenge.

In this paper, we investigate a simple approach for endowing policies with
memory, by augmenting the state space to include memory states that can be
written to by the policy. Na\"{i}vely using such a state representation with
standard policy search algorithms is quite challenging, because the algorithm
must simultaneously figure out how to use the memory states to choose the action
and how to store the right information into these states so that it can be
recalled later. The computations needed to make such decisions about memory
states require a complex, nonlinear policy structure. Such policies are
difficult to efficiently learn with model-free methods, while model-based
methods also require a model of the system dynamics, which can be difficult to
obtain \cite{dnp-spsr-13}. We show how the guided policy search algorithm can be
adapted to the task of training policies with internal memory.

In guided policy search, the policy is optimized using supervised learning
\cite{la-lnnpg-14,levine2015end}. The supervision is provided by using a simple
trajectory-centric reinforcement learning algorithm to individually solve the
task from a collection of fixed initial states. This trajectory-centric
``teacher'' resembles trajectory optimization. Since each teacher only needs to
solve the task from a single initial state, it is faced with a much easier
problem. The final policy is trained with supervised learning, which allows us
to use a nonlinear, high-dimensional representation for this final policy, such
as a multilayer neural network, in order to learn complex behaviors with good
generalization. A key component in guided policy search is adaptation between
the trajectories produced by the teacher and the final policy. This adaptation
ensures that, at convergence, the teacher does not take actions that the final
policy cannot reproduce. This is realized by an alternating optimization
procedure, which iteratively optimizes the policy to match each teacher, while
the teachers adapt to gradually match the behavior of the final policy.

To incorporate memory states into this method, we add the memory states to both
the trajectory-centric teacher and the final neural network policy. Since the
trajectories are adapted to the neural network policy, the teacher selects
memory states that will cause the neural network to take the right action,
essentially telling the network which information should be memorized to achieve
good performance. Because of this, the neural network only needs to learn how to
reproduce the memory states chosen by the teacher. The teacher effectively shows
the neural network which information needs to be written into the memory, and
the network need only figure out how to obtain this information from the
observations.

Our experimental results show that our method can be used to learn a variety of
continuous control tasks in manipulation and navigation settings. For example,
we show how memory states can allow a simulated robotic manipulator to remember
the target position for a peg insertion task, or place plates and bottles into
both vertical and horizontal slots, by remembering past sensory inputs from
contacts that allow it to determine the orientation of the opening. We also show
that using memory states with guided policy search outperforms other algorithms
and representations, including policies represented by LSTM neural networks and
alternative policy search methods that do not use a trajectory-centric teacher
to optimize the memory state values.

\section{Related Work}

While a complete survey of reinforcement learning methods for partially observed
problems is outside the scope of the paper, we highlight several relevant
research areas in this section. Discrete partially observed tasks have been
tackled using a variety of reinforcement learning and dynamic programming
methods \cite{ross2008online,shani2013survey}. While such methods have been
extended to small continuous spaces \cite{brunskill2008continuous}, they are
difficult to scale to the kinds of large state spaces found in most robotic
control tasks. In these domains, methods based on direct policy search are often
preferred, due to their ability to scale gracefully with task dimensionality
\cite{dnp-spsr-13}. While most policy search methods are concerned with reactive
policies, a number of methods have been proposed that augment the policy with
internal state, including methods based on finite state controllers
\cite{meuleau1999learning,toussaint2008hierarchical} and explicit memory states
that the policy can alter using memory storage actions
\cite{peshkin2001learning}. However, these methods have been evaluated only in
small or discrete settings. While our approach also supplies the policy with
internal memory states and explicit actions that can be used to alter that
state, our memory and storage actions are continuous, and our experiments show
that our method can scale to high-dimensional problems that are representative
of real-world robotic control tasks.

Taken together with their internal memory states, our policies can be regarded
as a type of recurrent neural network (RNN). Previous work has proposed training
RNN policies using likelihood ratio methods and backpropagation through time
\cite{wierstra2007solving}. However, this approach suffers from two challenges:
the first is that model-free likelihood ratio methods are difficult to scale to
policies with more than a few hundred parameters \cite{dnp-spsr-13}, which makes
it hard to apply the method to complex tasks that require flexible,
high-dimensional policy representations, and the second is that optimizing RNNs
with backpropagation through time is prone to vanishing and exploding gradients
\cite{pascanu2012difficulty}. While specialized RNN representations such as
LSTMs \cite{hochreiter1997lstm} or GRUs \cite{cho2014learning} can mitigate
these issues, we show that we can obtain better results by training the policy
to manipulate the memory states through explicit memory actions, without using
backpropagation through time. To that end, we extend the guided policy search
algorithm to train policies with memory states and memory actions.

The guided policy search algorithm used in this work is most similar to the
method proposed by Levine et al. \cite{lwa-lnnpg-15,levine2015end}. This
approach was proposed in the context of robotic control, and has been shown to
achieve good results with complex, high-dimensional feedforward neural network
policies. The central idea behind guided policy search is to decompose the
policy search problem into alternating trajectory optimization and supervised
learning phases, where trajectory optimization is used to find a solution to the
control problem and produce training data that is then used in the supervised
learning phase to train a nonlinear, high-dimensional policy. By training a
single policy from multiple trajectories, guided policy search can produce
complex policies that generalize effectively to a range of initial states.
Previous work has only applied guided policy search to training reactive
feedforward policies, since the algorithm assumes that the policy is Markovian.
We show the BADMM-based guided policy search method \cite{levine2015end} can be
extended to handle continuous memory states. The memory states are added to the
state of the system, and the policy is tasked both with choosing the action and
modifying the memory states. Although the resulting policy can be viewed as an
RNN, we do not need to perform backpropagation through time to train the
recurrent connections inside the policy. Instead, the memory states are
optimized by the trajectory optimization algorithm, which intuitively seeks to
set the memory states to values that will allow the policy to take the
appropriate action at each time step, and the policy then attempts to mimic this
behavior in the supervised learning phase.

\section{Background and Preliminaries}

The aim of our method is to control a partially observed system in order to
minimize the expectation of a cost function over the entire execution of a
policy $\policy_\params(\at|\obs_1,\dots,\obs_t)$, given by
$E_{\policy_\params}[\sum_{t=1}^T \cost(\st,\at)]$ in the finite-horizon
episodic setting. Here, $\st$ denotes the true state of the system, $\at$
denotes the action, $\ot$ denotes the observation, and $\cost(\st,\at)$ is the
cost function that specifies the task. For example, in the case of robotic
control, $\at$ might correspond to the torques at the robot's motors, $\st$
might be the configuration of the robot and its environment, including the
positions of task-relevant objects, and $\ot$ might be the readings from the
robot's sensors, such as joint encoders that provide the angles of the joints,
or even images from a camera. The policy
$\policy_\params(\at|\obs_1,\dots,\obs_t)$ specifies a distribution over actions
conditioned on the current and previous observations. This policy is
parameterized by $\params$. We are particularly concerned with tasks where the
current observation $\ot$ by itself is not sufficient for choosing a good action
$\at$, and the policy must integrate information from the past to succeed. Such
tasks require policies with internal state, which can be used to remember past
observations and act accordingly. To optimize policies with memory, we build on
the guided policy search algorithm presented by Levine et al.
\cite{levine2015end}, which we summarize briefly in this section. This algorithm
optimizes reactive policies of the form $\policy_\params(\at|\ot)$. We discuss
in Section~\ref{sec:rnngps} how it can be adapted to train policies with memory.

\subsection{Guided Policy Search}
\label{sec:gps}

Guided policy search is a policy optimization algorithm that transforms the
policy search task into a supervised learning problem, where supervision is
provided by a set of simple trajectory-centric controllers, denoted
$\trajdist_i(\at|\st)$, that are each optimized independently on separate
instances of the task, typically corresponding to different initial states.
There are two main benefits to this approach: the first is that, by requiring
each trajectory-centric controller to solve the task from only a specific
initial state, relatively simple controllers can be used that admit very
efficient reinforcement learning methods. The second benefit is that, since the
final policy is optimized with supervised learning methods, it can admit a
complex, highly expressive representation without concern for the usual
challenges associated with optimizing high-dimensional policies
\cite{dnp-spsr-13}. Intuitively, the purpose of the trajectory-centric
controllers is to determine how to solve the task from specific states, while
the purpose of the final policy is to generalize these controllers and succeed
from a variety of initial states. The partially observed variant of guided
policy search, which we build off of, takes this idea further, by also providing
a different input to the trajectory-centric controllers compared to the policy.
In this method, the trajectory-centric controllers are trained under full state
observation, while the policy is trained to mimic these controllers using only
the observations $\ot$ as input. This forces the policy to handle partial
observation, while keeping the task easy for the trajectory-centric controllers.
This type of instrumented setup is natural for many robotic tasks, where
training is done in a known laboratory setting, while the final policy must
succeed under a variety of uncontrolled conditions. However, this method does
not itself provide a way of handling internal memory.

\begin{algorithm}[tb]
    \caption{Partially observed guided policy search}
\label{alg:gps}
    \begin{algorithmic}[1]
        \FOR{iteration $k = 1$ to $K$}
            \STATE Run each $\trajdist_i(\at|\st)$ to generate samples $\{\traj_j\}$
            \STATE Fit local linear dynamics $\hat{\trajdist}_i(\state_{t+1}|\st,\at)$ around each $p_i(\at|\st)$ using $\{\traj_j\}$
            \FOR{inner iteration $l = 1$ to $L$}
                \STATE Optimize each $p_i(\at|\st)$ using fitted dynamics to minimize cost and match $\policy_\params(\at|\ot)$
                \STATE Optimize $\policy_{\params}(\at|\ot)$ to match all distributions $\trajdist_i(\at|\st)$ along each sample trajectory $\traj_j$
            \ENDFOR
        \ENDFOR
    \end{algorithmic}
\end{algorithm}

The partially observed guided policy search method is summarized in
Algorithm~\ref{alg:gps}. At each iteration of the algorithm, samples are
generated using each of the trajectory-centric controllers
$\trajdist_i(\at|\st)$.\footnote{We will drop the subscript $i$ from
$\trajdist_i(\at|\st)$ in the remainder of the paper for clarity of notation,
but all of the exposition extends trivially to the case of multiple
trajectory-centric controllers.} While a variety of representations for these
controllers are possible, linear-Gaussian controllers of the form
$\trajdist(\at|\st) = \gauss(\Kpol_t\st+\kpol_t,\ucovart)$ admit a particularly
efficient optimization procedure based on iterative refitting of local linear
dynamics \cite{la-lnnpg-14}. Once these dynamics are fitted, the algorithm takes
$L$ inner iterations (4 in our implementation). These iterations alternate
between optimizing each trajectory-centric controller $\trajdist(\at|\st)$ using
a variant of LQR under the fitted dynamics, and optimizing the policy
$\policy_\params(\at|\ot)$ to match the actions taken by the trajectory-centric
controllers at each observation $\ot^i$ encountered along the sampled
trajectories. The controllers are optimized to minimize their expected cost
$E_\trajdist[\cost(\traj)]$, as well as minimize their deviation from the
policy, measured in terms of KL-divergence. The policy is optimized to minimize
the KL-divergence from the controllers. This alternating optimization ensures
that the trajectory-centric controllers and the policy agree on the same
actions. In general, supervised learning is not guaranteed to produce good
long-term policies, since errors in fitting the action at each time step
accumulate over time \cite{rgb-rilsp-11}. Formally, the issue is that the policy
will not have the same state visitation frequency as the controllers it is
trained on. The alternating optimization addresses this by gradually forcing the
controllers and policy to agree. To ensure agreement, guided policy search uses
Lagrange multipliers on the means of the policy and the controllers, which are
updated every iteration. The full details of this method, including the
objectives for controller and policy optimization, are derived in previous work
\cite{levine2015end}.

\subsection{Trajectory-Centric Reinforcement Learning}
\label{sec:trajopt}

In guided policy search, the linear-Gaussian controllers $\trajdist(\at|\st)$
are optimized with respect to the cost $\cost(\st,\at)$, as well as an
additional term that penalizes deviation from the policy
$\policy_\params(\at|\st)$. This term consists of the KL-divergence between
$\trajdist(\at|\st)$ and $\policy_\params(\at|\st)$ with a weight $\admmrho_t$,
as well as a Lagrange multiplier $\lgmut$ on the mean action. Together, the cost
and the penalty form the following objective:
\begin{align}
L(\trajdist) = E_{\trajdist(\st,\at)}[ & \cost(\st,\at) - \at\tr\lgmut + \nonumber \\
& \admmrho_t \kl(\trajdist(\at|\st)\|\policy_\params(\at|\st)) ]. \nonumber
\end{align}
\noindent The linear-Gaussian controllers $\trajdist(\at|\st)$ can be optimized
in a variety of ways, including offline trajectory optimization methods with
known models \cite{lk-lcnnp-14} and trajectory-centric reinforcement learning
\cite{la-lnnpg-14}. We adopt the latter approach in this work, which we briefly
summarize in this section.

When the dynamics are locally smooth, a linear-Gaussian controller of the form
$\trajdist(\at|\st) = \gauss(\Kpol_t\st+\kpol_t,\ucovart)$ can be viewed as
inducing a mean trajectory with some linear feedback for stabilization. Hence,
we refer to the process of learning such controllers as trajectory-centric or,
more simply, as trajectory optimization. An efficient way to optimize these
controllers is to draw samples from the current $\trajdist(\at|\st)$, fit
time-varying linear-Gaussian dynamics to these samples of the form
$\hat{p}(\state_{t+1}|\st,\at) = \gauss(\fxt \st + \fut \at + \fct, \noise_t)$,
compute a local second-order Taylor expansion of the cost $\cost(\st,\at)$, and
then optimize the controller $\trajdist(\at|\st)$ using the LQR dynamic
programming algorithm. As described in previous work, this approach can achieve
sample-efficient learning for a variety of robotic manipulation skills
\cite{la-lnnpg-14,lwa-lnnpg-15}, but it requires an additional constraint to
ensure that the optimized controller remains in the region where the estimated
dynamics $\hat{p}(\state_{t+1}|\st,\at)$ are valid. This can be done by
constraining the KL-divergence between the new controller $\trajdist(\at|\st)$
and the previous controller $\bar{\trajdist}(\at|\st)$, which generated the
samples that were used to fit $\hat{p}(\state_{t+1}|\st,\at)$. The corresponding
optimization problem is given by
\[
\min_{\trajdist} L(\trajdist) \text{ s.t. } \kl(\trajdist(\traj)\|\bar{\trajdist}(\traj)) \leq \epsilon,
\]
where $\trajdist(\traj)$ is the trajectory distribution induced by
$\trajdist(\at|\st)$ and dynamics $\hat{\trajdist}(\state_{t+1}|\st,\at)$. Using
KL-divergence constraints for controller optimization has been proposed in a
number of previous works \cite{bs-cps-03,ps-rlmsp-08,pma-reps-10}, but in the
case of linear-Gaussian controllers, we can use a modified LQR algorithm to
solve this problem. We refer the reader to previous work for details
\cite{la-lnnpg-14}.

\subsection{Recurrent Neural Networks}

In order to avoid task-specific manual engineering of the policy class, guided
policy search is often used with general-purpose function approximators such as
large neural networks. One way to integrate memory into such policies is to use
recurrent neural networks (RNNs). Unlike feedforward networks, RNNs can maintain
a memory of past observations through their hidden states, which are propagated
forward in time according to the hidden state dynamics.

We can define an RNN with inputs $\ot$, outputs $\at$, and internal state
$\hidstate_t$ with two functions: an output function
\mbox{$\rnnout(\ot,\hidstate_t) = \at$} and a dynamics function
\mbox{$\rnndyn(\ot,\hidstate_t) = \hidstate_{t+1}$}. In practice, $\rnnout$ and
$\rnndyn$ might share some parameters, but viewing them as separate functions
will make it convenient for us to compare standard RNNs with our memory states,
which we describe in the next section.

RNNs are typically trained by viewing them as one large neural network and
computing the gradient of the parameters with respect to the loss by using
backpropagation through time. However, learning long-term temporal dynamics is
still very difficult for RNNs, since backpropagation through time can lead to
vanishing and exploding gradients. Many solutions have been proposed for these
issues. One popular solution consists of altering the architecture of the
network to make optimization easier, with the LSTM architecture being
particularly popular. We therefore evaluate such an architecture as the baseline
in our experiments in Section~\ref{sec:experiments}.

Guided policy search provides us with an easier and more effective method for
training such policies, by including the hidden states (referred to as memory
states for clarity) directly into the state of the dynamical system. This avoids
the need for using backpropagation through time, and instead uses trajectory
optimization to optimize the memory state values. This approach, which we
describe in detail below, achieves significantly better results in our
experiments, and has a number of appealing computational benefits.

\section{Memory States}

\begin{figure}
    \setlength{\unitlength}{0.5\columnwidth}
    \begin{picture}(1.99,1.15) \linethickness{0.5pt}
    \put(0.025, -1.3){\includegraphics[width=\linewidth]{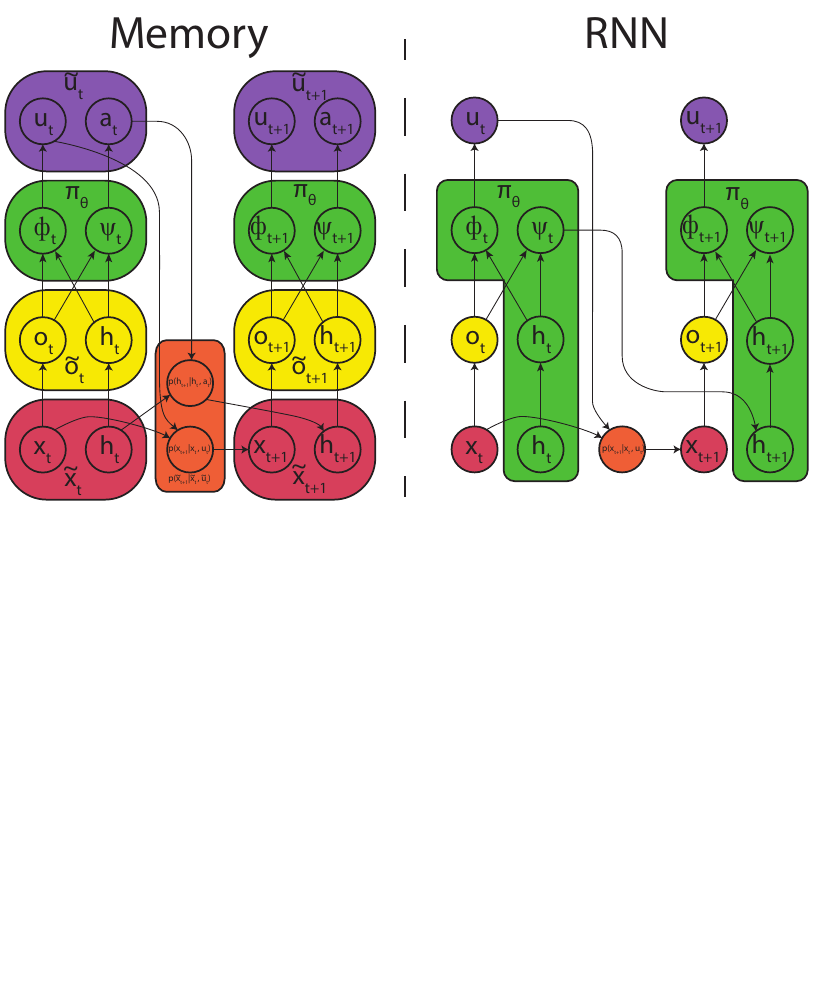}}
    \put(0.16,1.02){(a)}
    \put(1.3,1.02){(b)}
    \end{picture}
    \caption{
        Diagram comparing memory states (a) and recurrent neural networks (b).
        In the case of memory states, both $\rnnout$ and $\rnndyn$ are
        incorporated into the policy, rather than treating $\rnndyn$ as a hidden
        state dynamics function. This makes the apparent dynamics of the memory
        states independent of the network parameters, making it easy to apply
        guided policy search.  Note that the computational units $\rnnout$ and
        $\rnndyn$ are identical in both cases, the only difference is in whether
        $\hidstate_t$ is considered to be part of the policy or part of the
        state.
        \label{fig:memdiagram}
    }
\end{figure}

Instead of directly optimizing RNNs with backpropagation through time, we
consider a different method for integrating memory into the policy. In our
approach, the memory states $\vt$ are directly concatenated to the physical
state of the system $\st$ and the observation $\ot$, to produce an augmented
state $\tst$ and augmented observation $\tot$, and the action is also augmented
to include memory writing actions $\vat$ to produce an augmented action $\tat$:
\[
\tst = \left[\begin{matrix}\st\\\vt\end{matrix}\right] \hspace{0.3in} \tot = \left[\begin{matrix}\ot\\\vt\end{matrix}\right] \hspace{0.3in} \tat = \left[\begin{matrix}\at\\\vat\end{matrix}\right].
\]
The dynamics then factorize according to
\[
p(\tilde{\state}_{t+1}|\tst,\tat) = p(\state_{t+1}|\st,\at)p(\hidstate_{t+1}|\vt,\vat),
\]
where there are various ways to choose $p(\hidstate_{t+1}|\vt,\vat)$ depending
on the semantics of $\vat$. In this work, we define
\mbox{$p(\hidstate_{t+1}|\vt,\vat) = \gauss(\vt + \vat, \sigma^2\mathbf{I})$},
where $\sigma^2$ is chosen to be a small constant ($10^{-6}$ in our
implementation) to ensure that the trajectory distributions remains
well-conditioned.

Training a policy $\policy_\params(\tat|\tot)$ on this augmented dynamical
system produces a policy that, in principle, can use the memory actions $\vat$
to write to the memory states $\vt$. In practice, the particular choice of
policy optimization algorithm makes a significant difference in how well the
policy can utilize the memory states, since there is no guidance on how they
should be used, aside from overall task performance. In the next section, we
describe the relationship between these policies and RNNs, and in
Section~\ref{sec:rnngps}, we will describe how the guided policy search
algorithm can train policies that effectively utilize memory states. In
Section~\ref{sec:experiments} we will show that this approach can produce
effective policies that succeed on a range of simulated manipulation and
navigation tasks that require memory.

\subsection{Comparison of Memory States and RNNs}

Policies of the form $\policy_\params(\tat|\tst)$, that use states and actions
augmented with memory, form RNNs when combined with the memory state dynamics
\mbox{$p(\hidstate_{t+1}|\vt,\vat)$}. In fact, these RNNs are in general
stochastic, though we use linear-Gaussian memory dynamics
\mbox{$p(\hidstate_{t+1}|\vt,\vat)$} with a small variance, as described in the
previous section, which makes them effectively deterministic in our
implementation. Note, however, that $\policy_\params(\tat|\tst)$ by itself is
not recurrent. This distinction is illustrated in Figure~\ref{fig:memdiagram},
which compares the structure of an RNN with output function $\rnnout(\ot,\vt)$
and dynamics function $\rnndyn(\ot,\vt)$ to the policy
$\policy_\params(\tat|\tst)$ with memory state dynamics
\mbox{$p(\hidstate_{t+1}|\vt,\vat)$}.

Aside from the stochastic aspects of $\policy_\params(\tat|\tst)$ and
\mbox{$p(\hidstate_{t+1}|\vt,\vat)$}, which become neglible as the variance of
both functions goes to zero, the relationship between this structure and the RNN
is that $\policy_\params(\tat|\tst)$ contains both $\rnnout(\ot,\vt)$ and
$\rnndyn(\ot,\vt)$. For example, when $\policy_\params(\tat|\tst)$ is Gaussian,
with a mean that depends on $\tst$ and a constant covariance, and
\mbox{$p(\hidstate_{t+1}|\vt,\vat)$} has the form in the previous section, we
have
\[
E_{\policy_\params(\tat|\tst)}[\tat | \tst] = \left[ \begin{matrix} \rnnout(\ot,\vt) \\ \rnndyn(\ot,\vt) - \vt \end{matrix} \right],
\]
\noindent so that the policy outputs the action $\rnnout(\ot,\vt)$ and the next
hidden state is $\rnndyn(\ot,\vt)$. Thus, we see that any RNN can be encoded as
a non-recurrent policy with memory states. Furthermore, when
$\policy_\params(\tat|\tst)$ and \mbox{$p(\hidstate_{t+1}|\vt,\vat)$} have
non-negligible stochasticity, memory states can be used to encode stochastic
recurrent networks. While the variance of the policies in our experiments is
independent of $\tat$, it would be straightforward to extend our method with
more complex stochastic policies.

\subsection{Guided Policy Search with Memory States}
\label{sec:rnngps}

Memory states can in principle be combined with any policy search algorithm. In
fact, prior work has proposed using discrete memory with storage actions
\cite{peshkin2001learning}. However, for high-dimensional, continuous tasks, the
logic required to choose which information to store and recall and when can
become quite complex. This necessitates the use of powerful, expressive function
approximators with hundreds or even thousands of parameters to represent
$\policy_\params(\tat|\tst)$, which are generally very difficult to train with
standard policy search techniques \cite{dnp-spsr-13}. Guided policy search has
previously been shown to be effective at learning these types of policies, and
in this section we describe how guided policy search can be adapted to handle
memory states.

Since the memory states and memory writing actions are simply appended to the
observation and action vectors, the supervised learning procedure for the policy
remains identical, and the policy is automatically trained to use the memory
actions to mimic the pattern of memory activations optimized by the
trajectory-centric ``teacher'' algorithms. The trajectory-centric teacher
optimizes linear-Gaussian controllers $\trajdist(\tat|\tst)$ that control both
the physical and memory states, essentially choosing the memory that the policy
needs to have in order to take the right action. This happens automatically,
because guided policy search adds a term to the cost function that penalizes
deviation from the policy $\policy_\params(\tat|\tst)$ in terms of
KL-divergence. As described Section~\ref{sec:trajopt}, this penalty takes the
form of a KL-divergence and a linear Lagrange multiplier term.

We make a small modification to the trajectory-centric teacher algorithm to
account for the particularly simple structure of the memory states. This
modification also helps to make the algorithm scalable to larger memory state
dimensionalities. Optimizing the linear-Gaussian controllers
$\trajdist(\tat|\tst)$ requires estimating the dynamics
$\hat{p}(\tstate_{t+1}|\tst,\tat) = \gauss(\tilde{f}_{\tstate t}\tst +
\tilde{f}_{\taction t}\tat + \tilde{f}_c,\noise_t)$, which we do by using linear
regression with a Gaussian mixture model prior, as described in previous work
\cite{la-lnnpg-14}. This approach is highly sample efficient, but we can make it
even more efficient in the case of memory states by exploiting our knowledge of
their dynamics. To that end, the dynamics are fitted according to
\[
\tilde{f}_{\tstate t} = \left[\begin{matrix} \fxt & 0 \\ 0 & \mathbf{I} \end{matrix}\right] \hspace{0.25in} \tilde{f}_{\taction t} = \left[\begin{matrix} \fut \\ \mathbf{I} \end{matrix}\right] \hspace{0.25in} \tilde{f}_c = \left[\begin{matrix} \fct \\ 0 \end{matrix} \right],
\]
where $\fxt$, $\fut$, and $\fct$ are estimates of the dynamics of the physical
system, computed from the samples in the same way as in prior work
\cite{la-lnnpg-14}. Aside from this modification, the guided policy search
algorithm we employ follows Algorithm~\ref{alg:gps}.

\section{Experimental Results}
\label{sec:experiments}

\begin{figure*}
    \setlength{\unitlength}{\columnwidth}
    \begin{picture}(1.99,0.7) \linethickness{0.5pt}
        \put(0.12, 0.66){Peg sorting}
        \put(0.00, 0.35){\includegraphics[width=0.4\columnwidth]{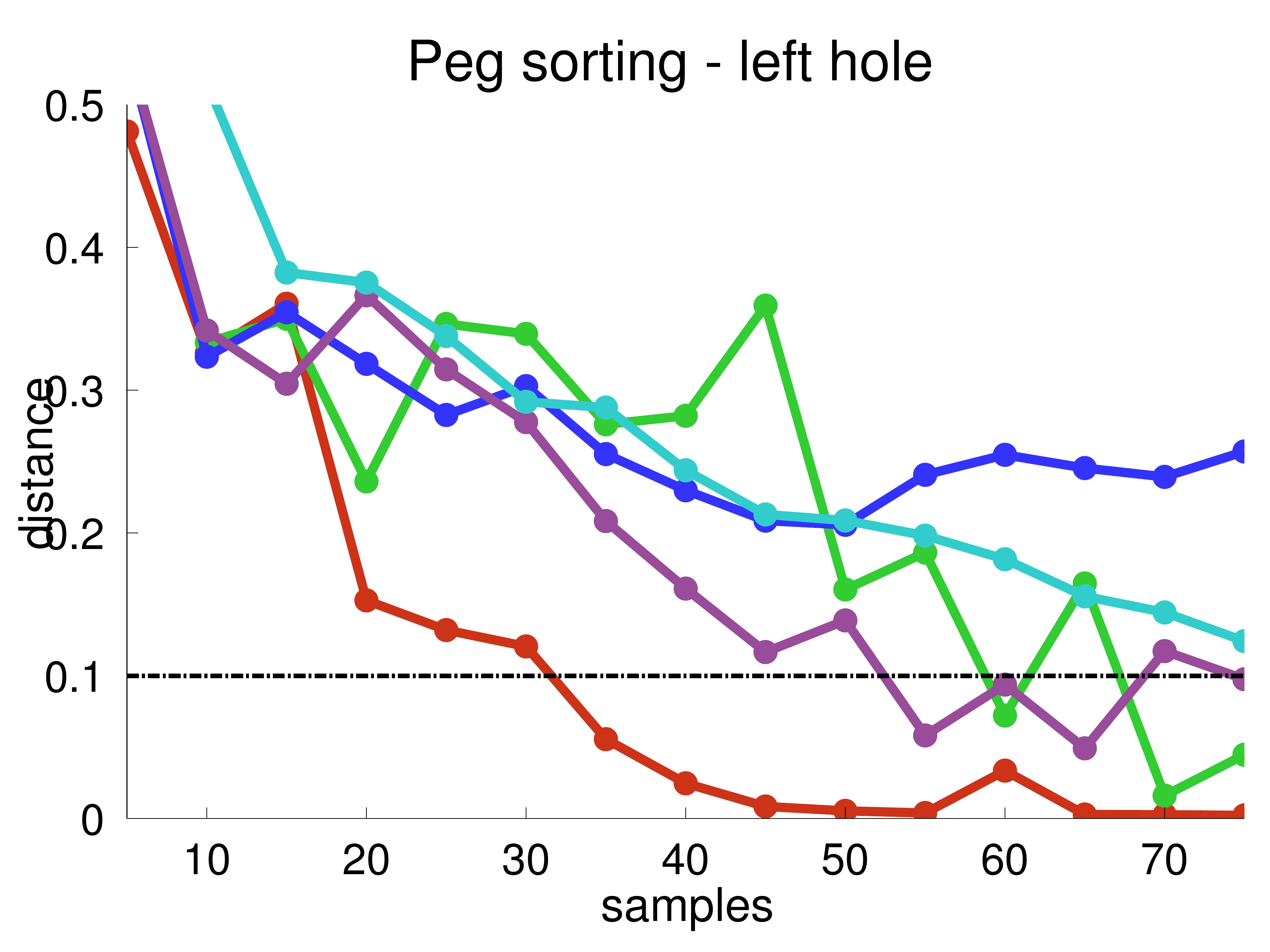}}
        \put(0.00, 0.03){\includegraphics[width=0.4\columnwidth]{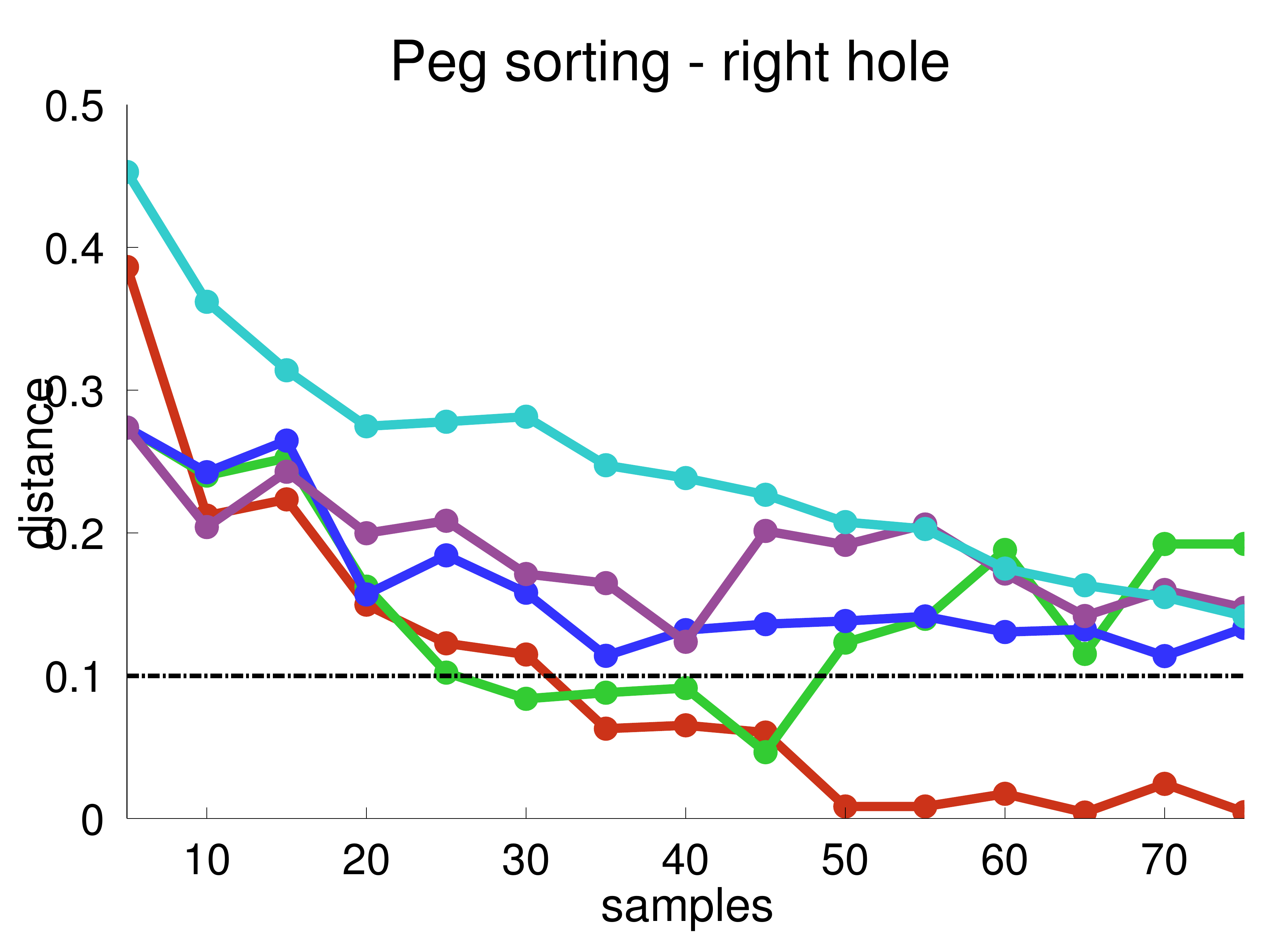}}
        \put(0.43, 0.03){\line(0,1){0.67}}
        \put(0.67, 0.66){Bottle and plate task}
        \put(0.45, 0.35){\includegraphics[width=0.38\columnwidth]{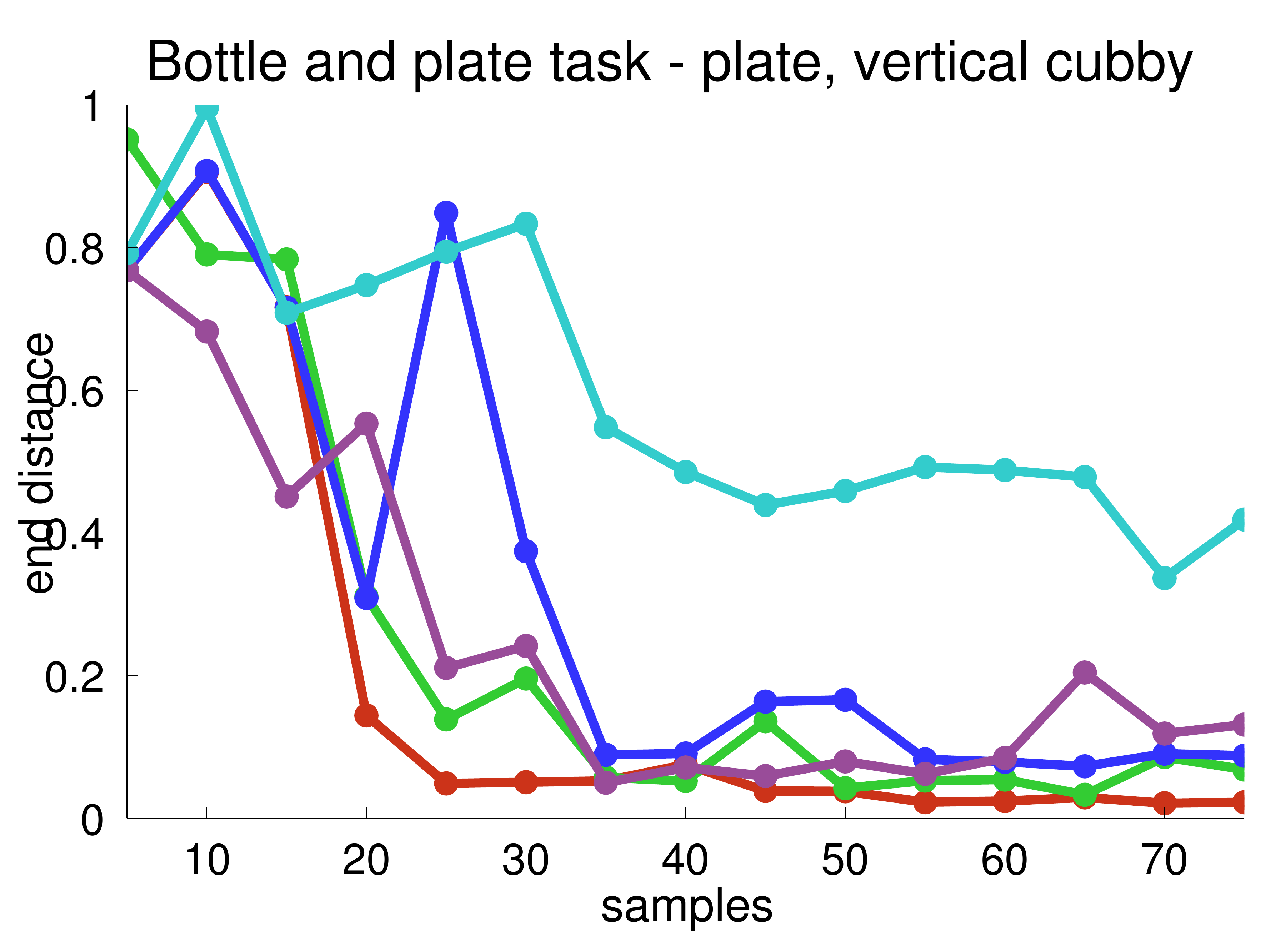}}
        \put(0.85, 0.35){\includegraphics[width=0.38\columnwidth]{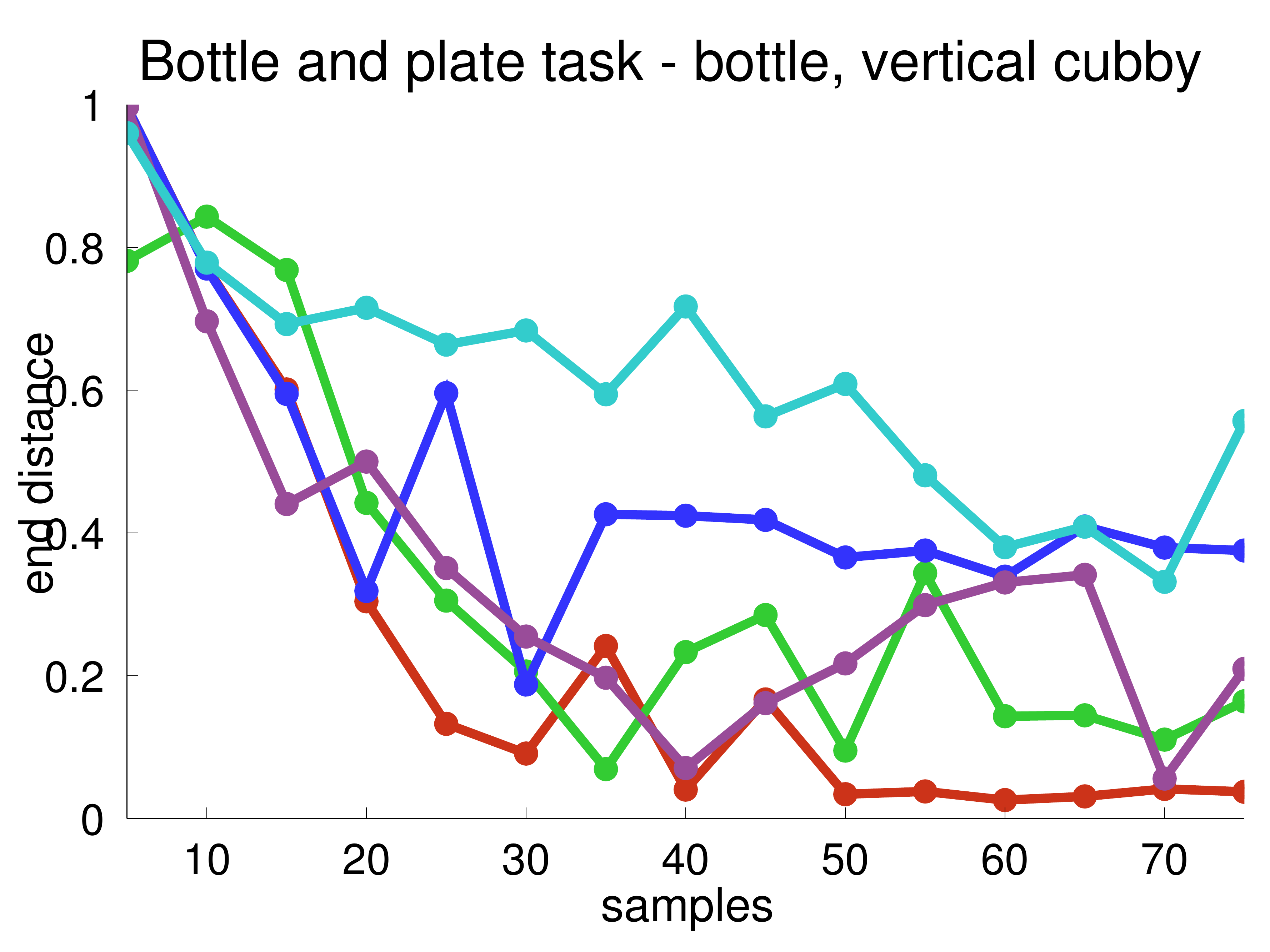}}
        \put(0.45, 0.03){\includegraphics[width=0.38\columnwidth]{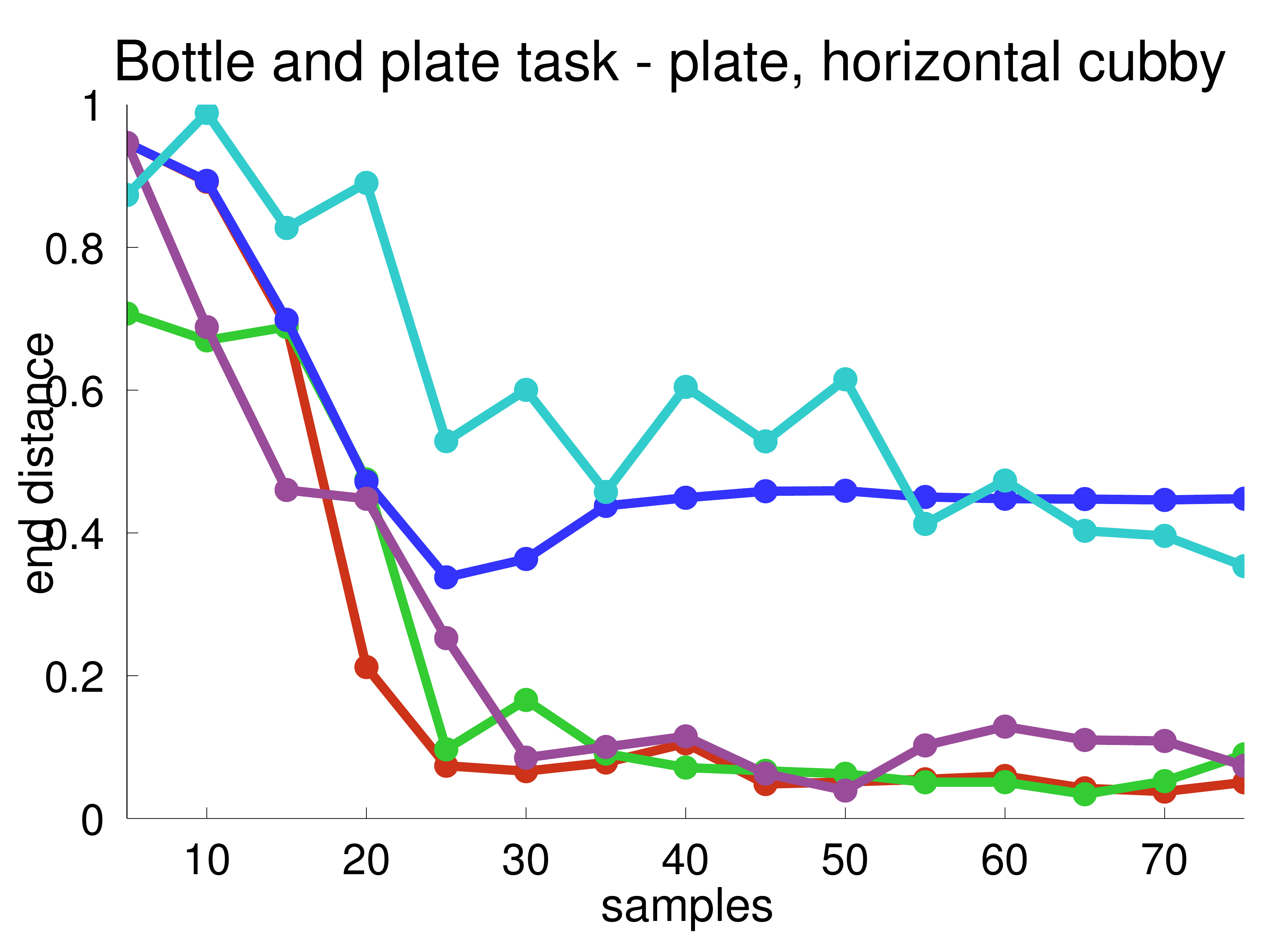}}
        \put(0.85, 0.03){\includegraphics[width=0.38\columnwidth]{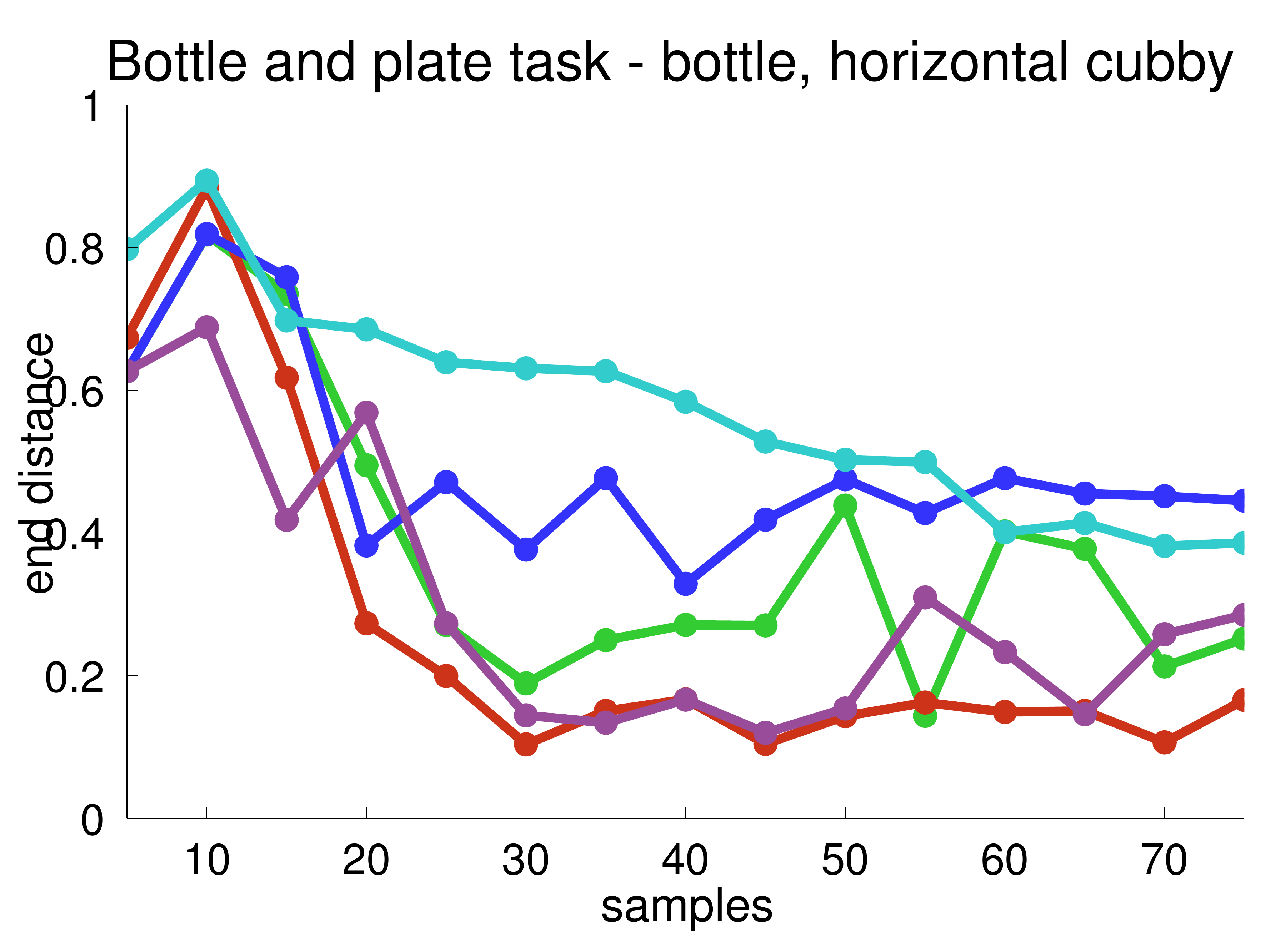}}
        \put(1.27, 0.03){\line(0,1){0.67}}
        \put(1.55, 0.66){2D navigation}
        \put(1.29, 0.35){\includegraphics[width=0.38\columnwidth]{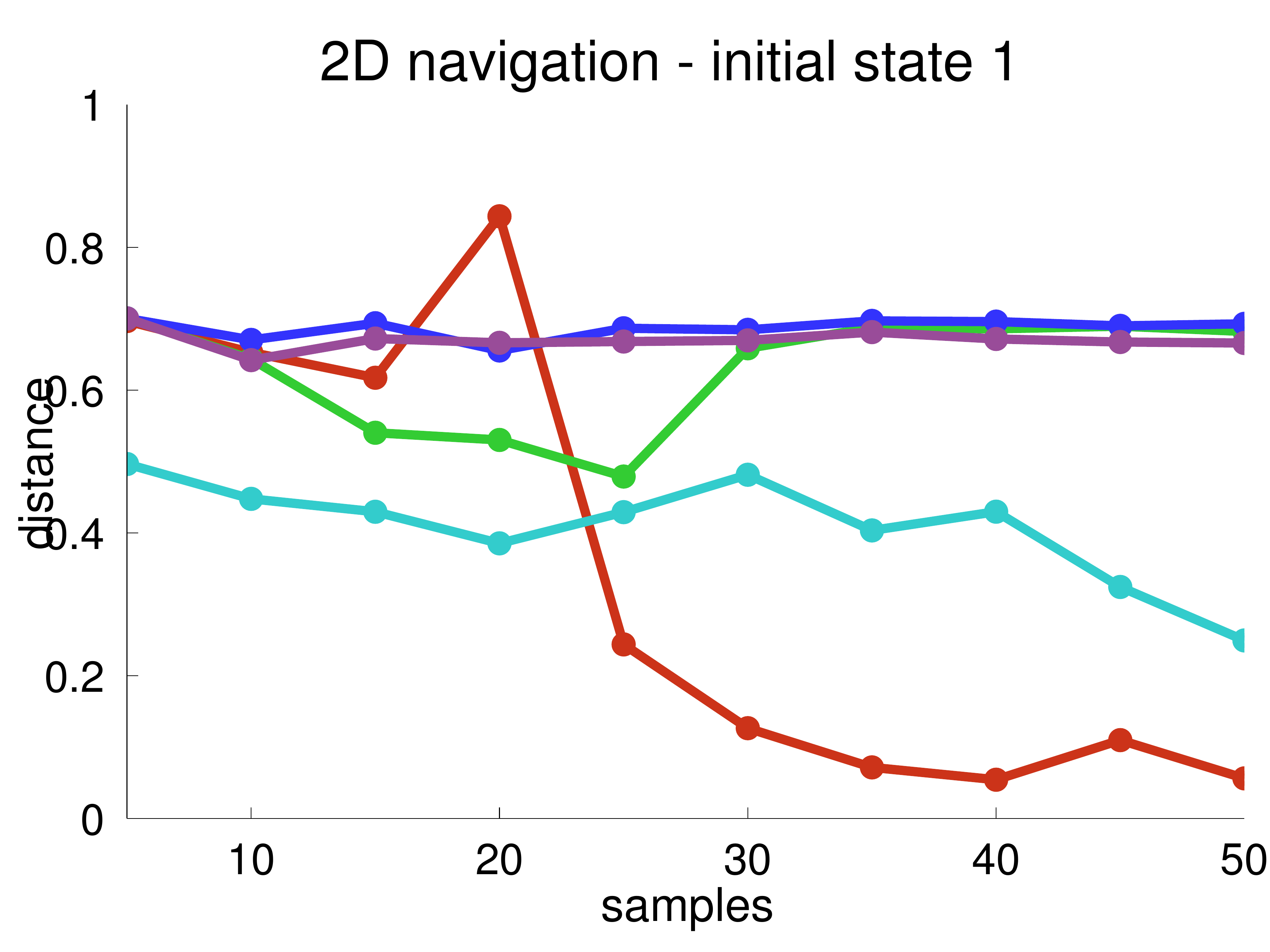}}
        \put(1.68, 0.35){\includegraphics[width=0.38\columnwidth]{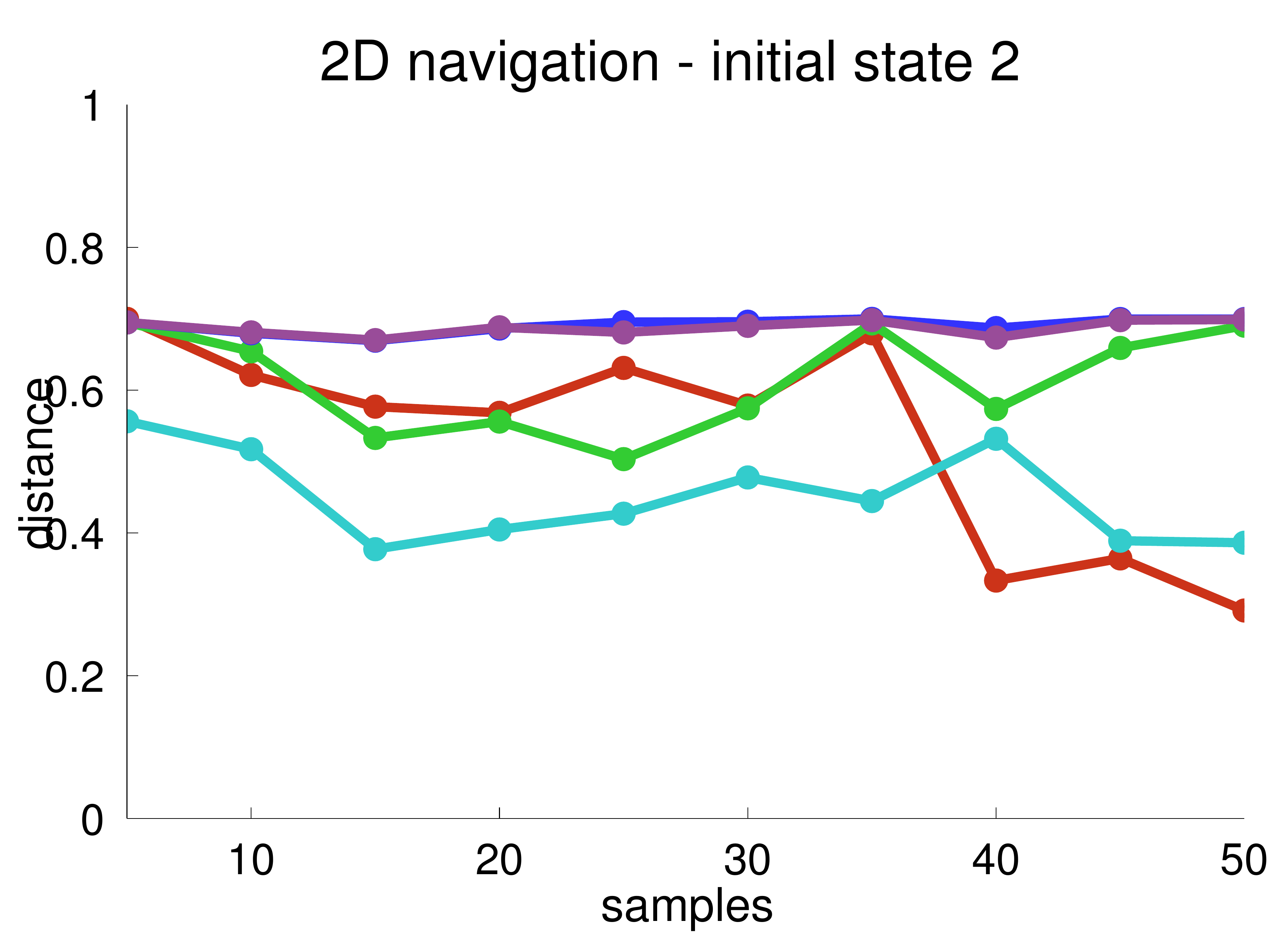}}
        \put(1.29, 0.03){\includegraphics[width=0.38\columnwidth]{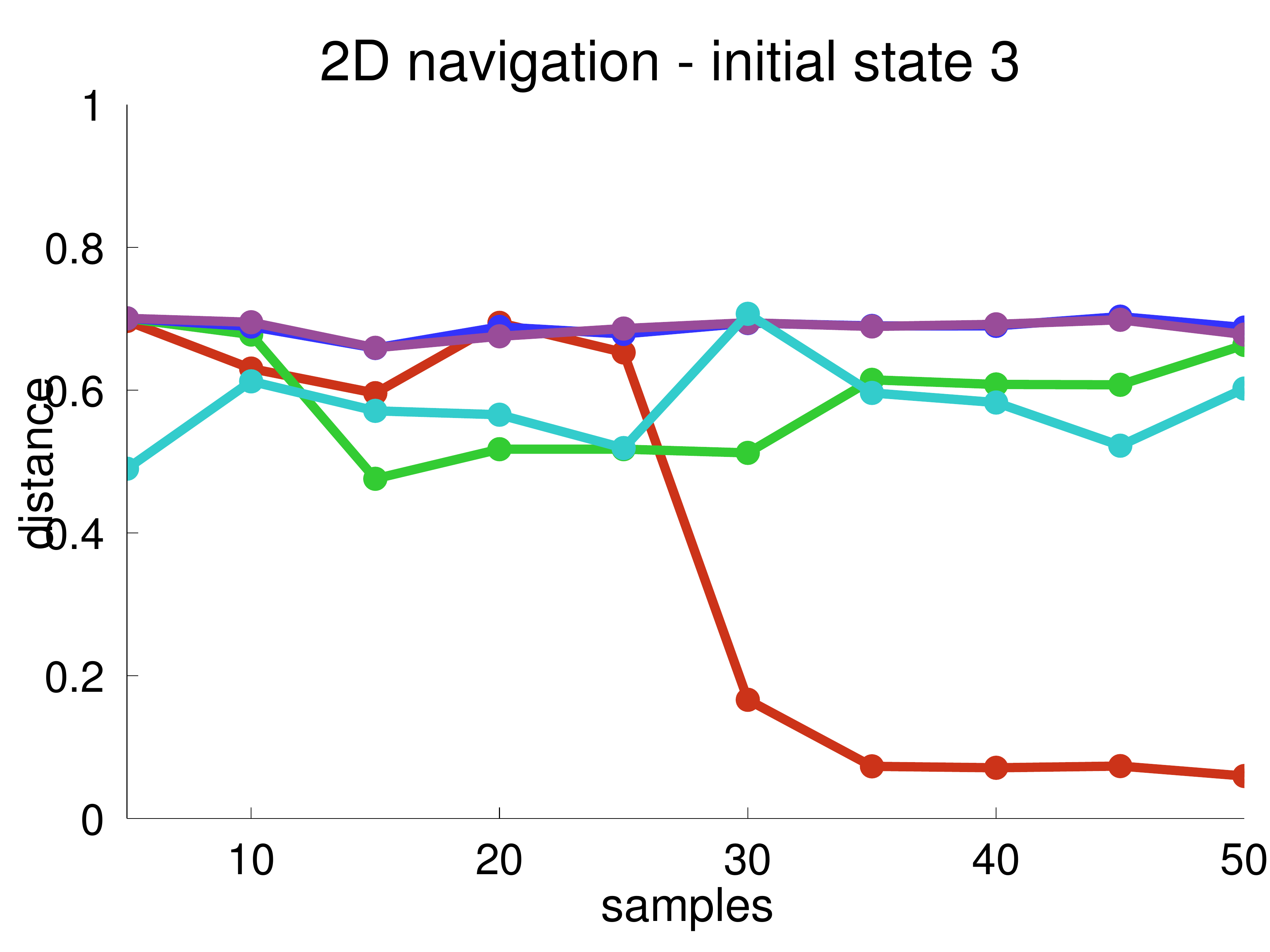}}
        \put(1.68, 0.03){\includegraphics[width=0.38\columnwidth]{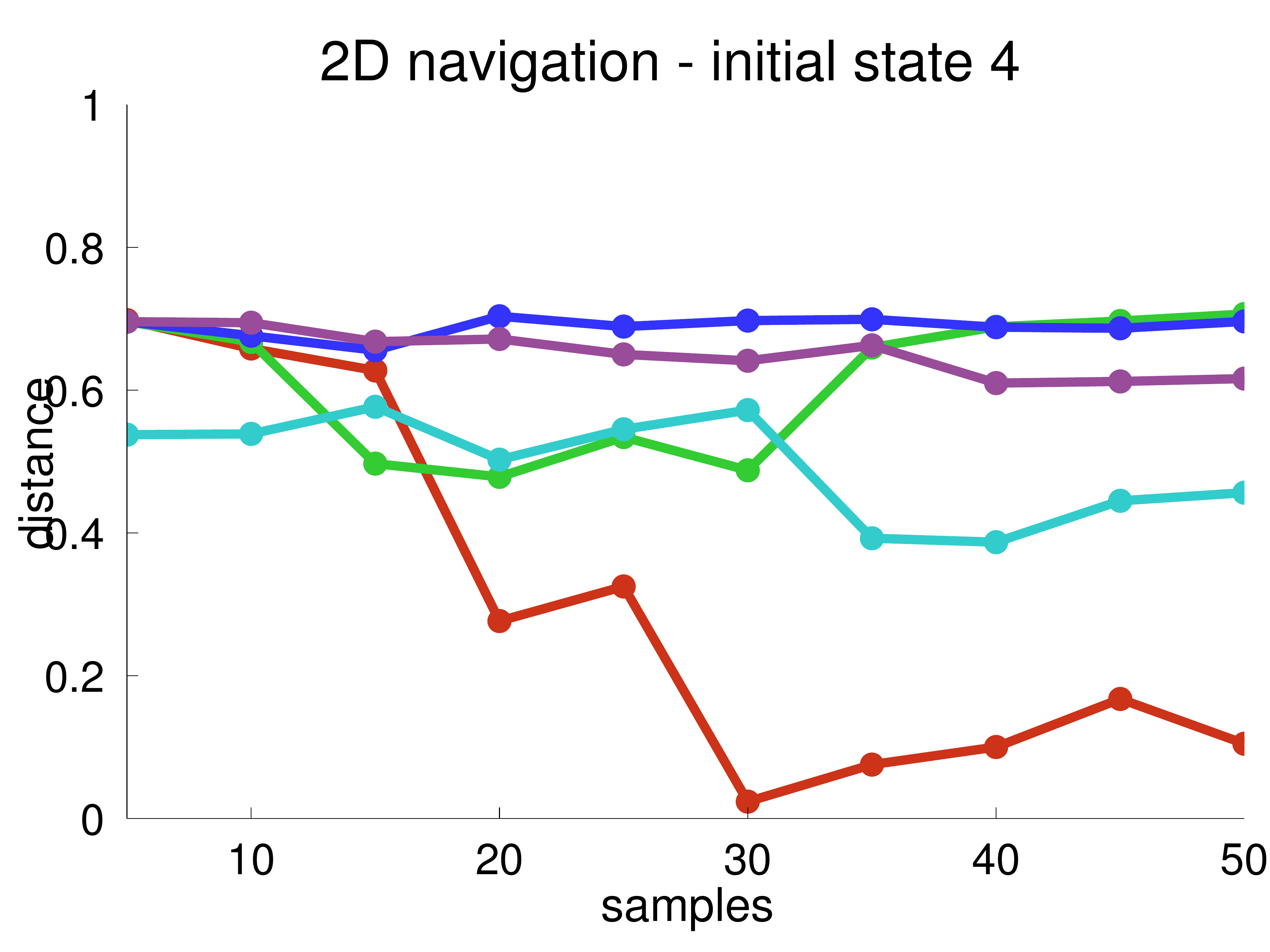}}
        \put(0.04, -0.03){\includegraphics[width=2.0\columnwidth]{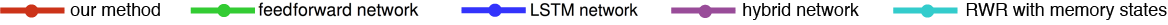}}
    \end{picture}
    \caption{
        Plots of the distance to the target in terms of the number of samples.
        For each method, the different lines show the distance for a different
        target position under the same policy. For the peg sorting and bottle
        and plate tasks, we plot the distance from the peg, while the retrieval
        task shows the distance between the retrieved object and the agent's
        starting position. In the sorting task, distances greater than the black
        dotted line correspond to failed trajectories. Note that a successful
        policy must succeed on \emph{all} of the conditions for the task. Our
        policy with memory states is able to successfully solve each of the
        tasks, while the alternative architectures and methods fail on at least
        one condition for each of the tasks.
        \vspace{-0.15in}
        \label{fig:results}
    }
\end{figure*}
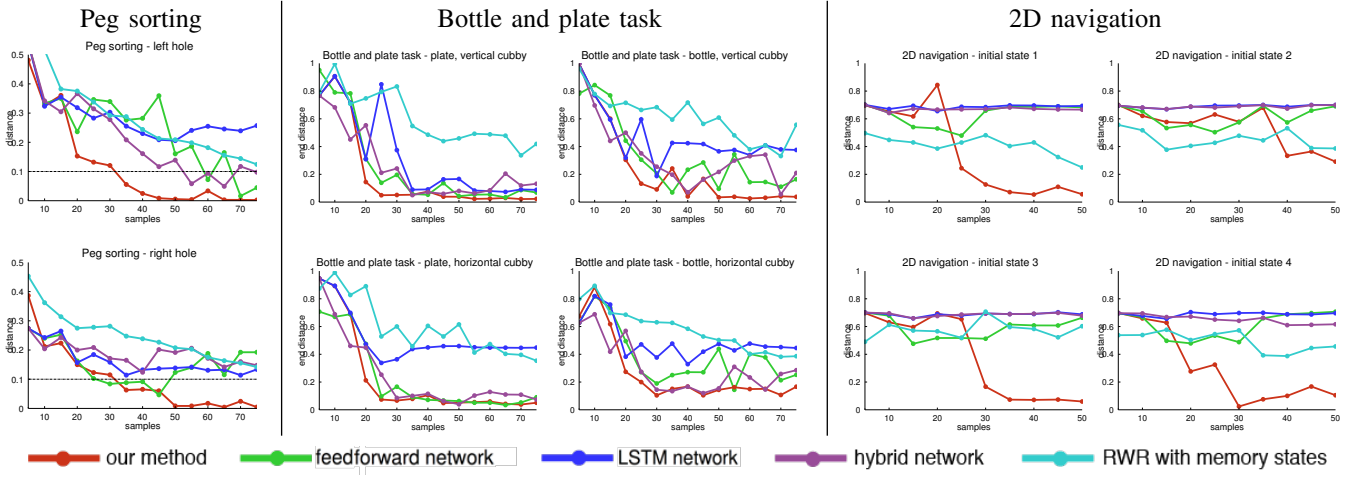

We evaluated our approach on a simple 2D navigation task, as well as two
high-dimensional simulated tasks involving robotic manipulation, and compared it
to alternative policy architectures and training methods. The aim of these
experiments was to answer the following questions:
\begin{enumerate}
  \item Can guided policy search with memory states solve complex tasks that
      require memory?
  \item How do memory states compare with more standard RNN policies?
  \item Does guided policy search make it easier to train policies with memory
      states, compared to alternative policy search methods?
\end{enumerate}
The purpose of the 2D navigation task is to provide a platform for comparing the
various methods and representations that is physically simple, but requires
memory to succeed. The purpose of the more complex manipulation tasks is to
evaluate the methods on a task that requires handling complex dynamics and
high-dimensional state, and therefore requires a policy that both has memory and
can learn complex control functions.

\subsection{Representations and Methods}
\label{sec:baselines}

For the manipulation tasks, we used a neural network policy with two hidden
layers of 40 rectified linear units (Relu) of the form $z = \max(a,0)$, while
the navigation task used a single hidden layer with 10 units. The manipulation
policies used 7-dimensional memory states, while the navigation policies used
4-dimensional memory states.

\begin{figure}
    \setlength{\unitlength}{0.5\columnwidth}
    \begin{picture}(1.99,1.25) \linethickness{0.5pt}
        \put(0.075, -1.1){\includegraphics[width=\linewidth]{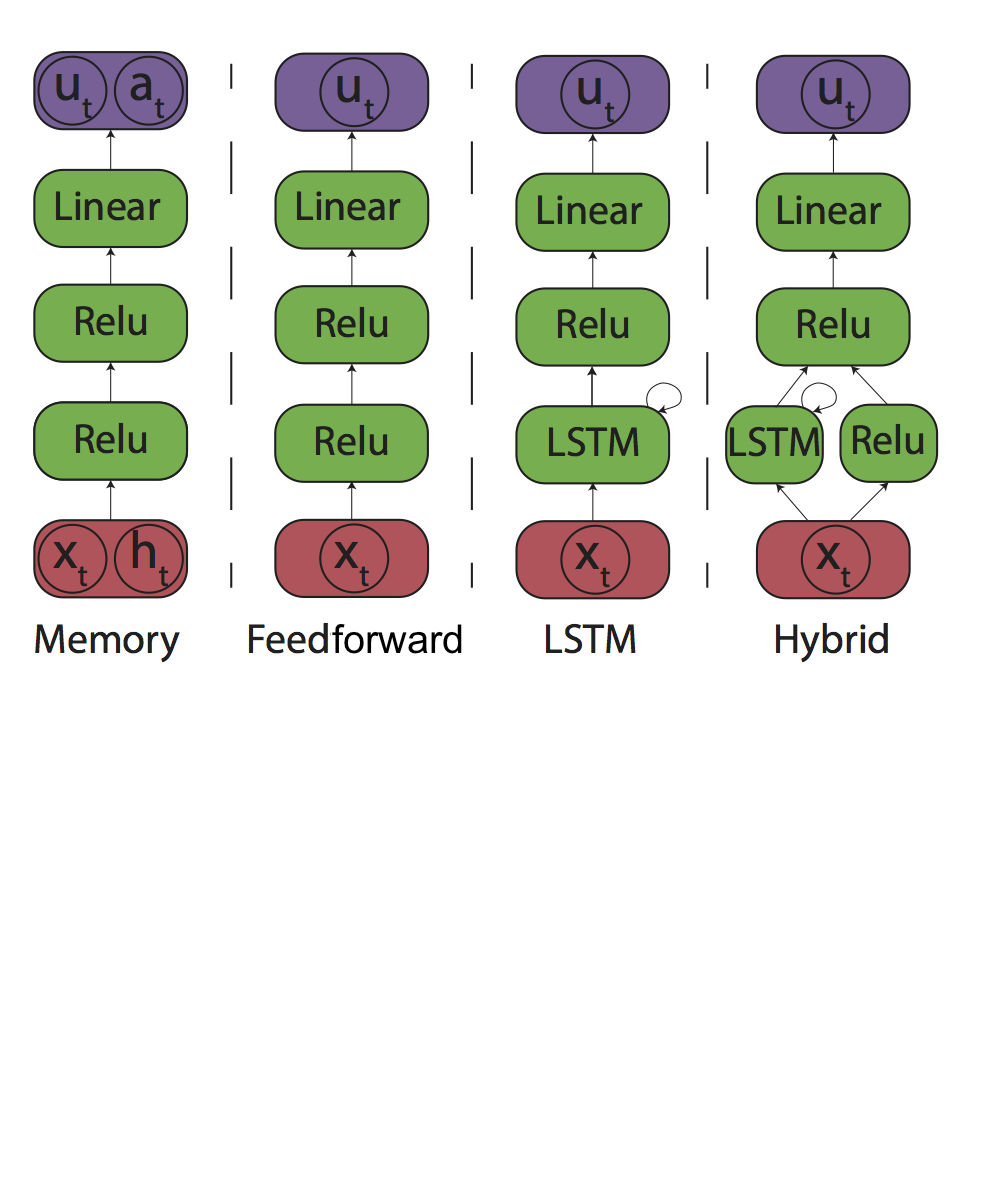}}
    \end{picture}
    \caption{
        Illustrations of the architecture used by our policy, as well as the
        alternative architectures.
        \label{fig:arch}
    }
    \vspace{-0.2in}
\end{figure}

In addition to guided policy search with memory states, we evaluated three
alternative policy representations, all trained with guided policy search, as
well as an alternative optimization algorithm. The first of these
representations was a feedforward neural network, without memory, but with the
same number of units as the memory states policy. This type of policy was used
with guided policy search in previous work \cite{la-lnnpg-14}, but it cannot
learn tasks that require preserving information from previous time steps. The
second representation was a recurrent network with LSTM units, which have
previously been shown to achieve good results for long-term memorization tasks
\cite{hochreiter1997lstm} and have recently become the architecture of choice
for recurrent networks \cite{sutskever2014sequence}. The last representation was
a hybrid network that consisted of both a feedforward branch and a recurrent
LSTM branch at the first layer. We constructed this hybrid representation after
observing that the standard LSTM policies often performed worse than the purely
feedforward network. Illustrations of each of the architectures are shown in
Figure~\ref{fig:arch}.

Besides guided policy search, we evaluated the memory states approach with the
reward-weighted regression (RWR) algorithm \cite{ps-aenac-07,kb-lmpr-09}. In
previous work, we observed that this method performed well on tasks with
high-dimensional policy representations \cite{la-lnnpg-14}, making it a good
baseline method for our tasks. We found that on all tasks, RWR achieved better
results with a linear policy than with a neural network, so all reported RWR
results use a linear parameterization.

\subsection{Tasks}

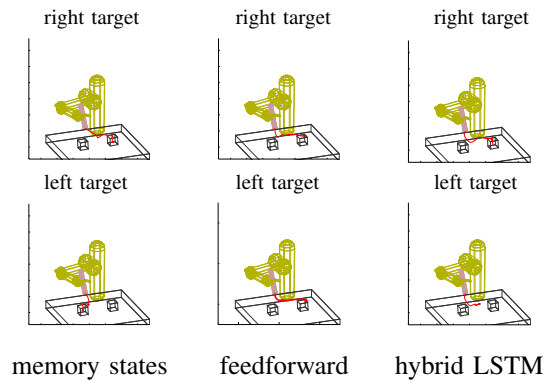
\begin{figure}
    \setlength{\unitlength}{0.5\columnwidth}
    \begin{picture}(1.99,1.1) \linethickness{0.5pt}
        \put(0.14, 0.60){\includegraphics[width=0.24\columnwidth]{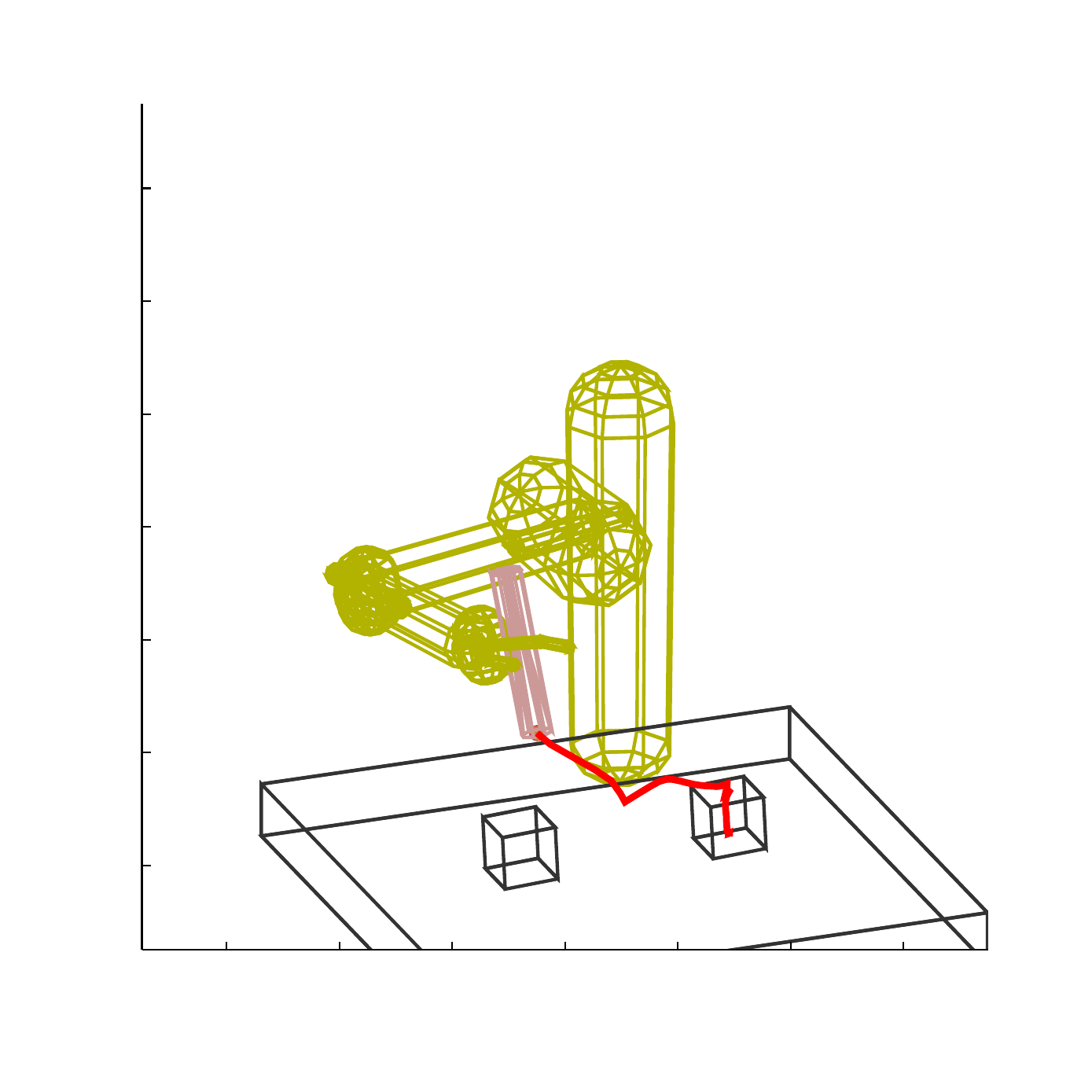}}
        \put(0.14, 0.09){\includegraphics[width=0.24\columnwidth]{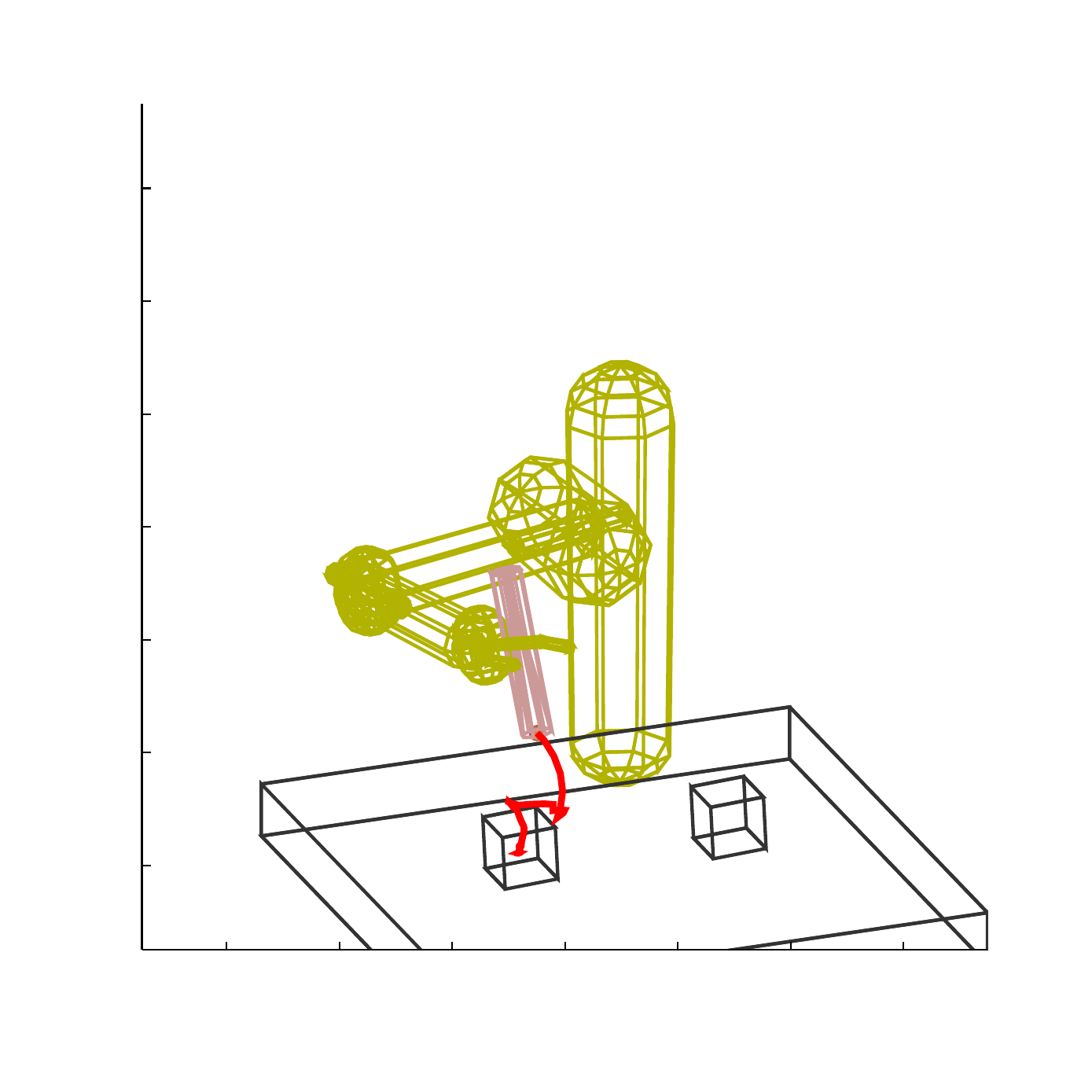}}
        \put(0.150, 0.00){memory states}
        \put(0.724, 0.60){\includegraphics[width=0.24\columnwidth]{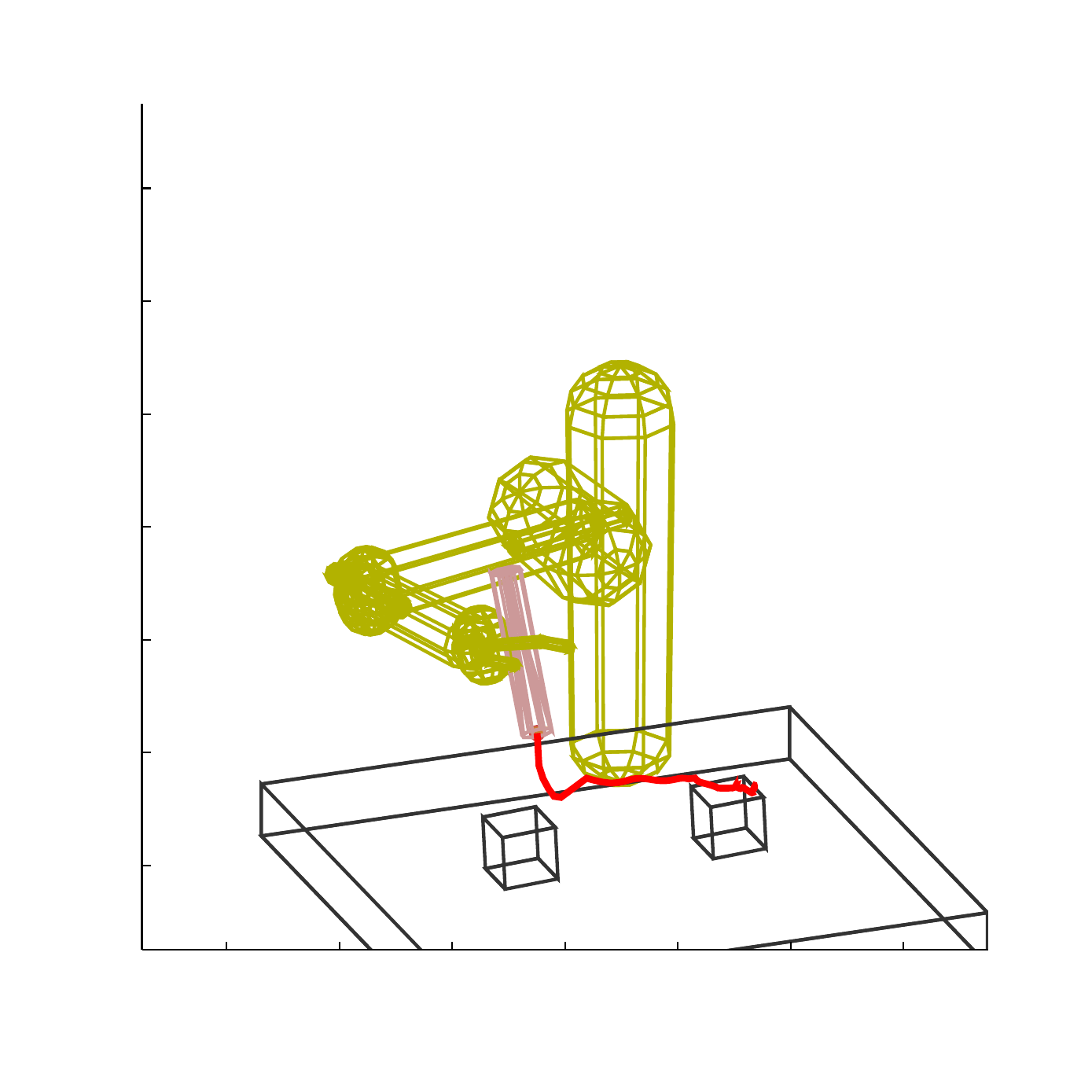}}
        \put(0.784, 0.15){\includegraphics[width=0.19\columnwidth]{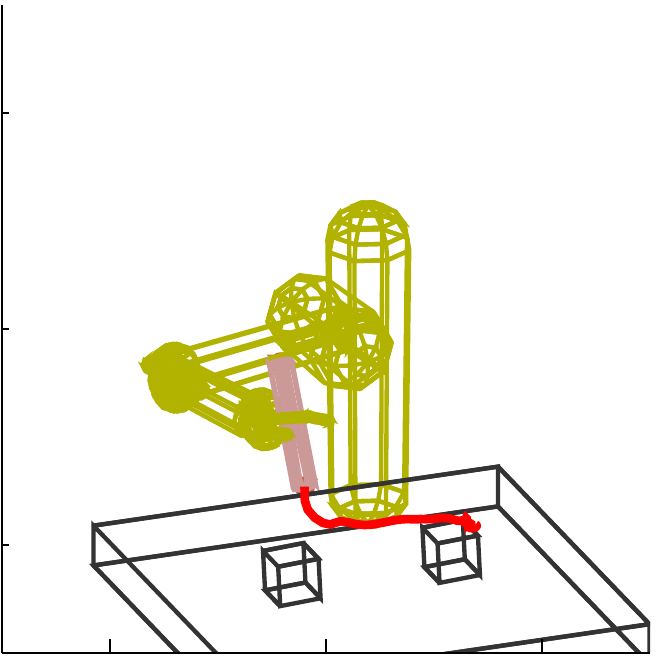}}
        \put(0.790, 0.00){feedforward}
        \put(1.369, 0.65){\includegraphics[width=0.19\columnwidth]{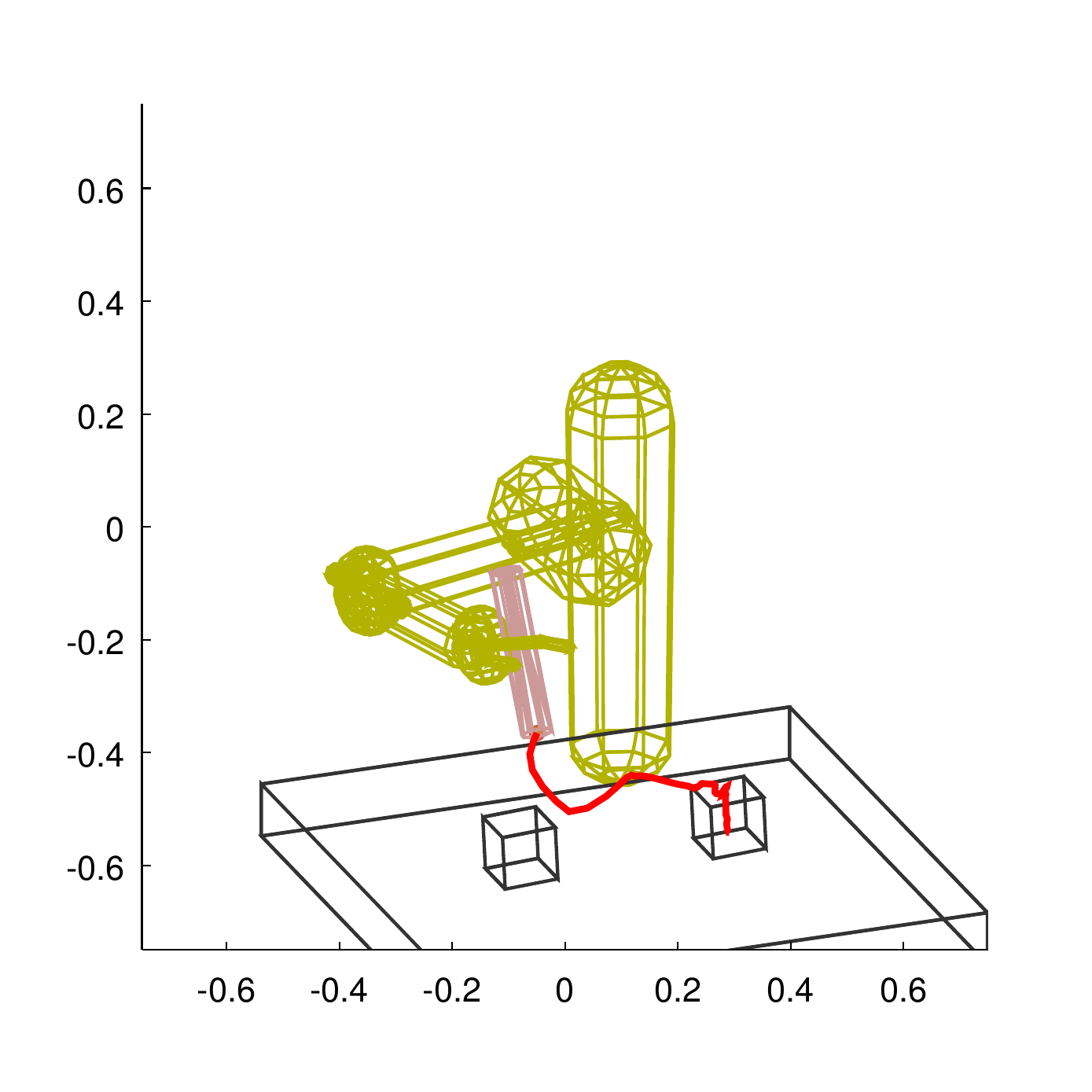}}
        \put(1.369, 0.15){\includegraphics[width=0.19\columnwidth]{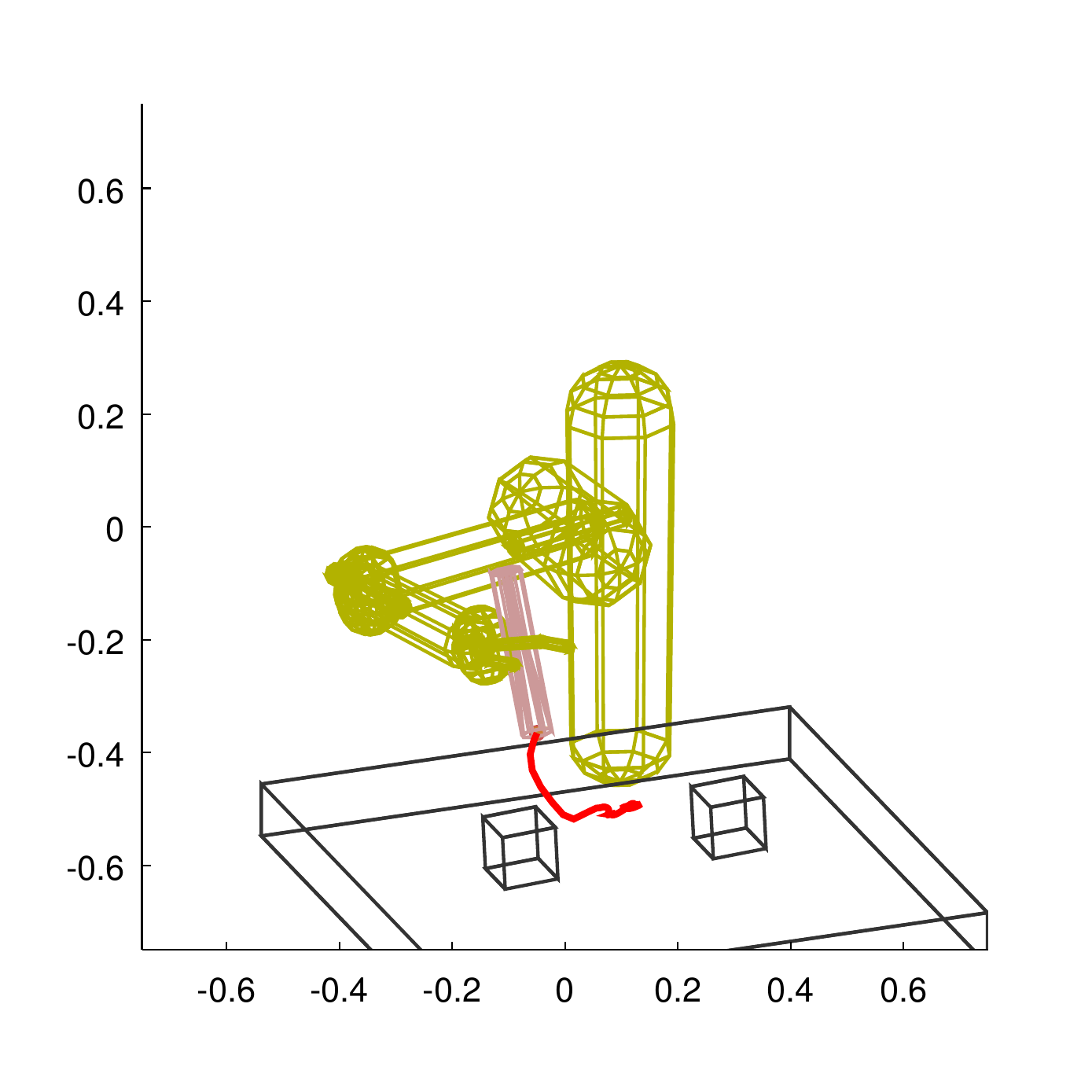}}
        \put(1.330, 0.00){hybrid LSTM}
    {\footnotesize
        \put(0.250, 1.08){right target}
        \put(0.250, 0.57){left target}
    }
    {\footnotesize
        \put(0.84, 1.08){right target}
        \put(0.84, 0.57){left target}
    }
    {\footnotesize
        \put(1.45, 1.08){right target}
        \put(1.45, 0.57){left target}
    }
    \end{picture}
    \caption{
        Sample trajectories for the peg sorting task for our our method, the
        feedforward network, and the hybrid LSTM network. Note that our method
        chooses the right target for both conditions.
        \label{fig:pegsort}
    }
    \vspace{-0.15in}
\end{figure}

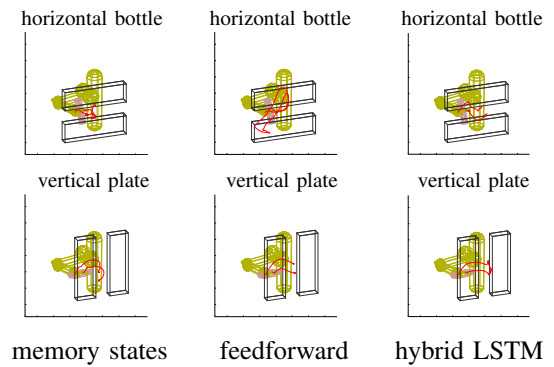
\begin{figure}
    \setlength{\unitlength}{0.5\columnwidth}
    \begin{picture}(1.99,1.1) \linethickness{0.5pt}
        \put(0.19, 0.60){\includegraphics[width=0.19\columnwidth]{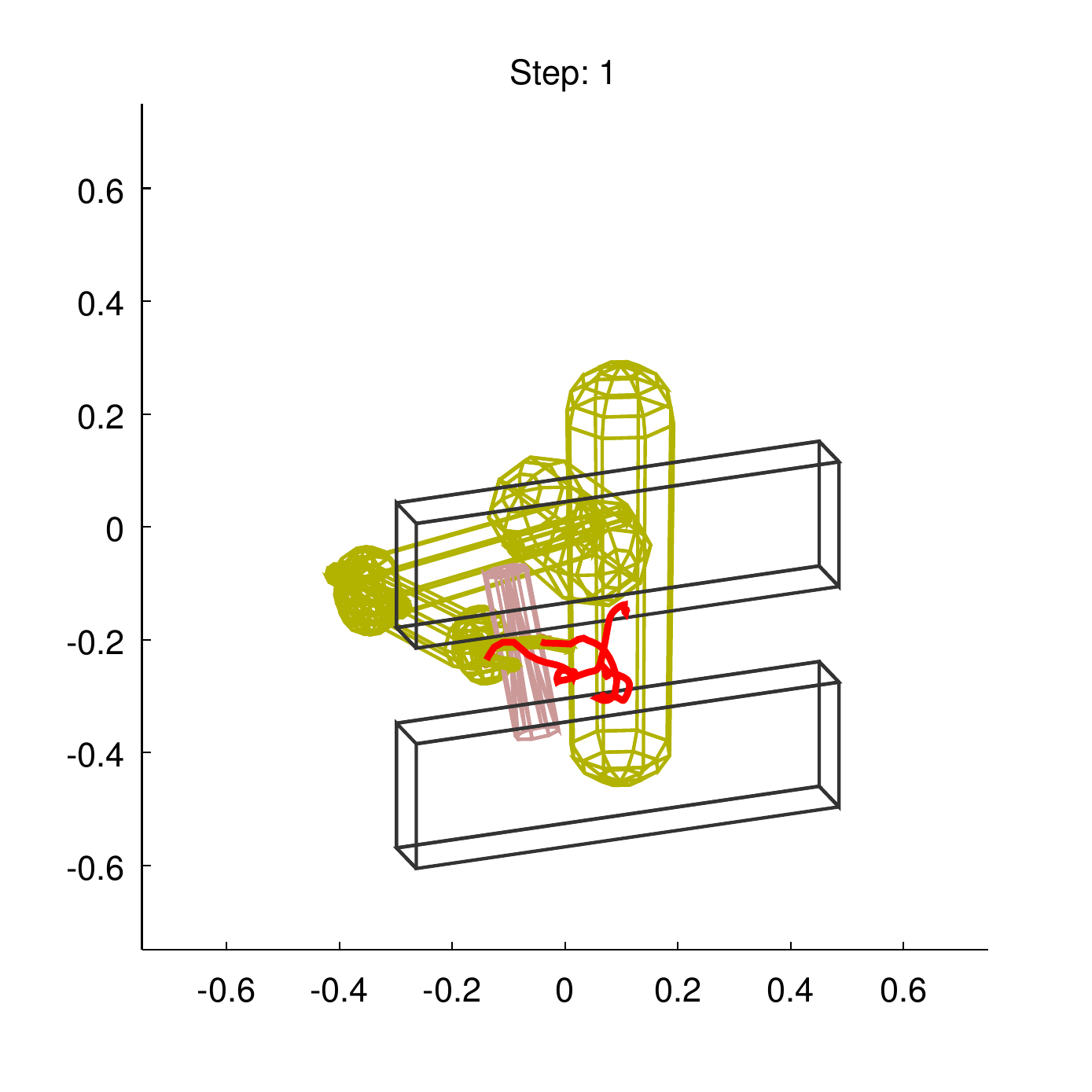}}
        \put(0.19, 0.10){\includegraphics[width=0.19\columnwidth]{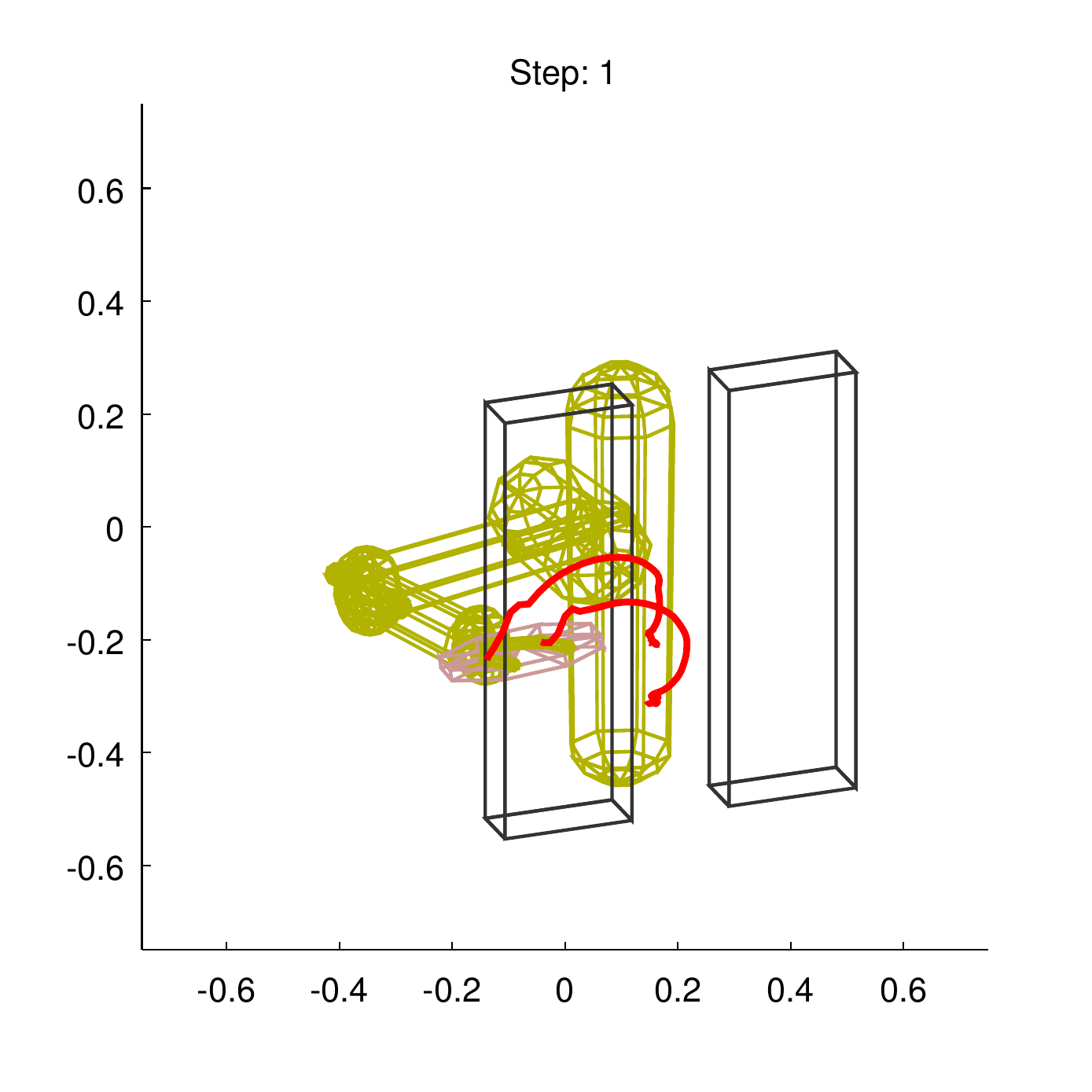}}
        \put(0.150, -0.03){memory states}
        \put(0.774, 0.60){\includegraphics[width=0.19\columnwidth]{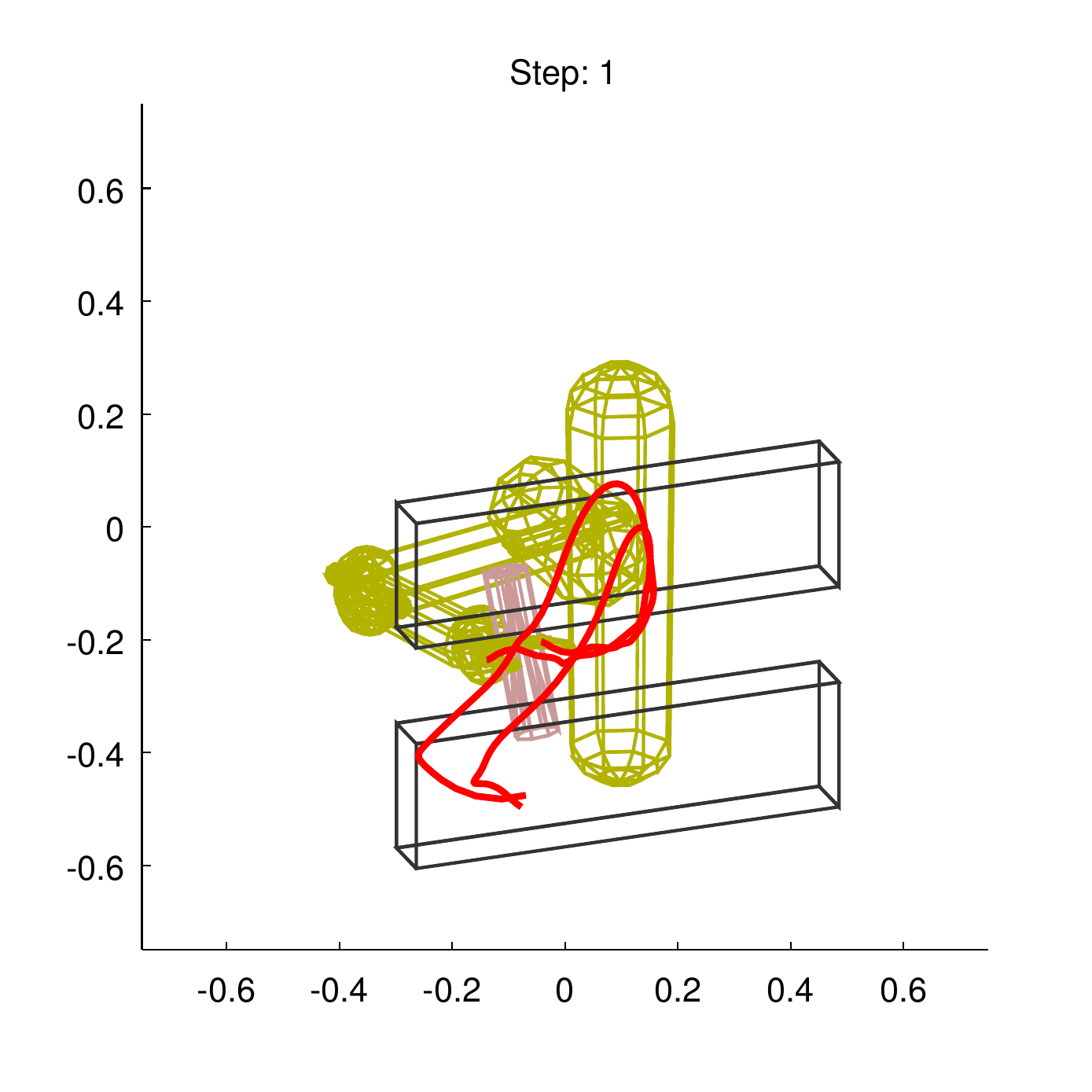}}
        \put(0.774, 0.10){\includegraphics[width=0.19\columnwidth]{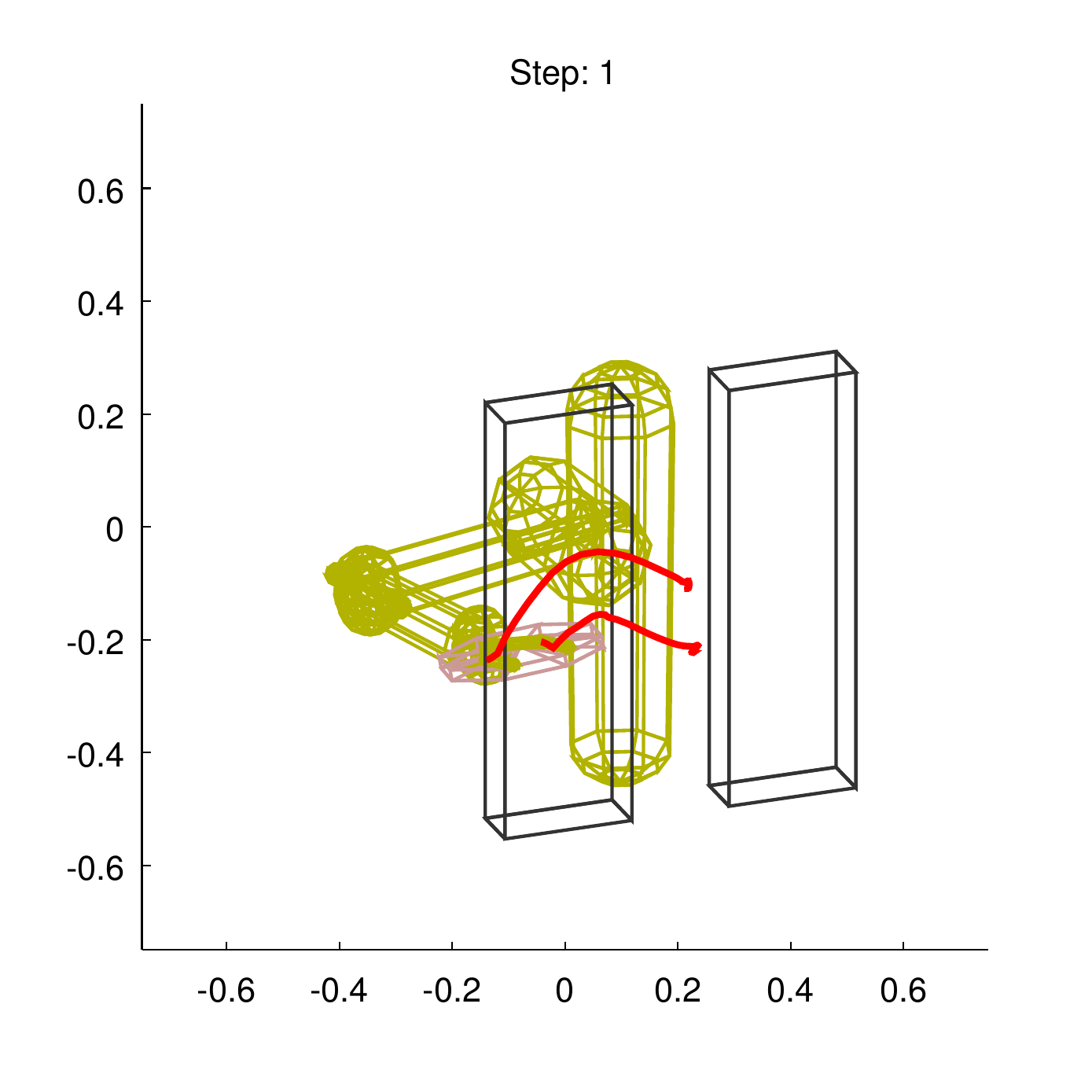}}
        \put(0.790, -0.03){feedforward}
        \put(1.369, 0.60){\includegraphics[width=0.19\columnwidth]{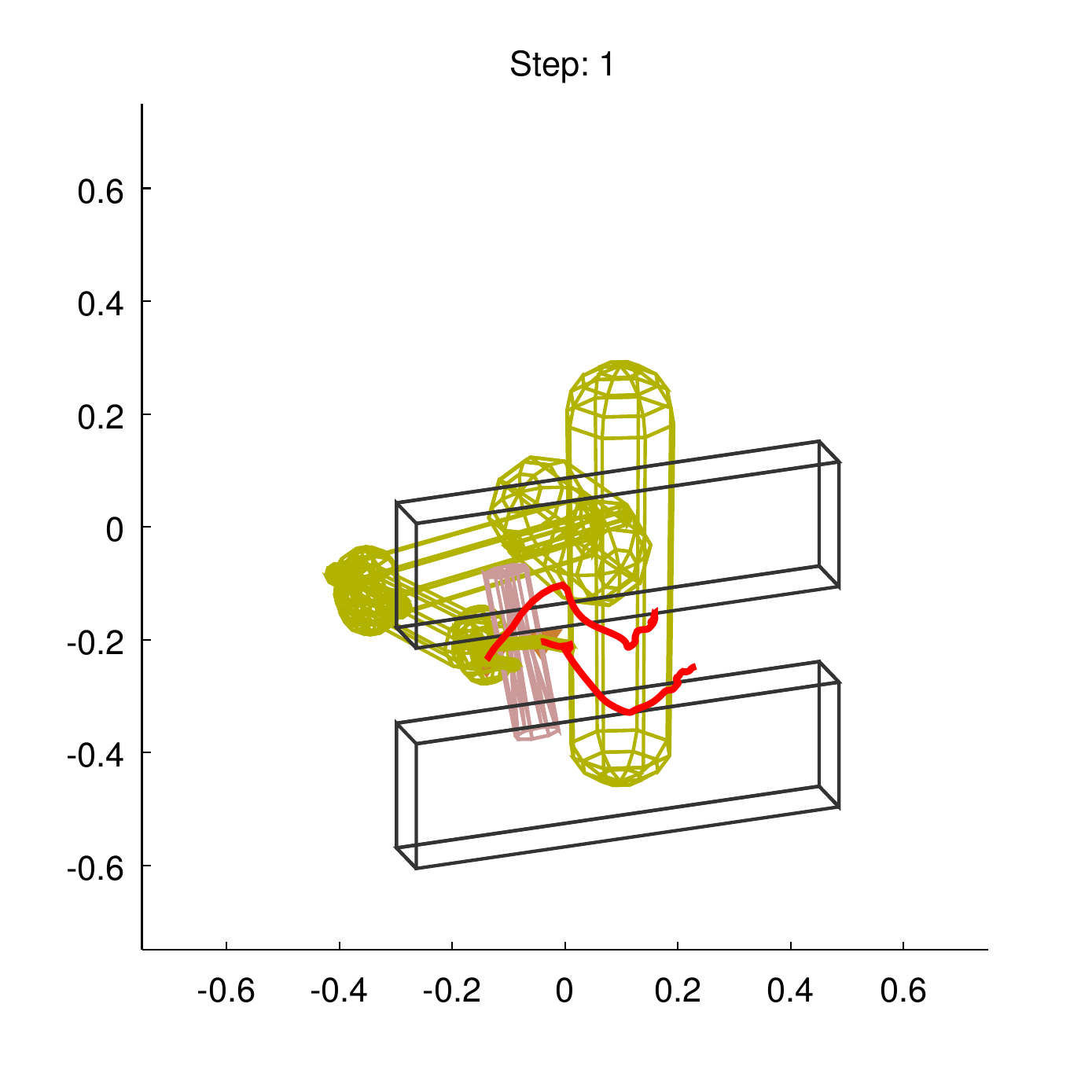}}
        \put(1.369, 0.10){\includegraphics[width=0.19\columnwidth]{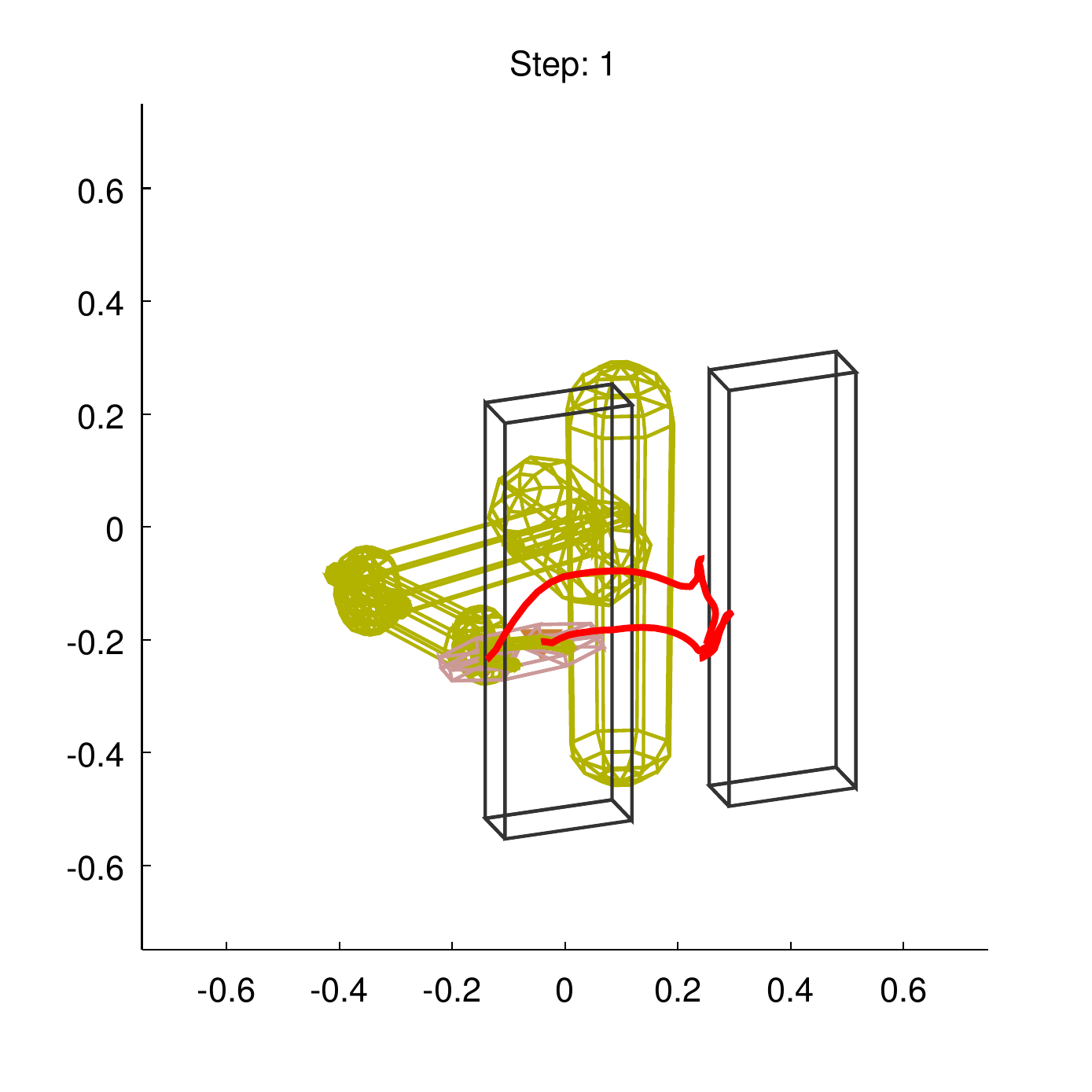}}
        \put(1.330, -0.03){hybrid LSTM}
    {\footnotesize
        \put(0.18, 1.0){horizontal bottle}
        \put(0.23, 0.5){vertical plate}
    }
    {\footnotesize
        \put(0.76, 1.0){horizontal bottle}
        \put(0.81, 0.5){vertical plate}
    }
    {\footnotesize
        \put(1.35, 1.0){horizontal bottle}
        \put(1.4, 0.5){vertical plate}
    }
    \end{picture}
    \caption{
        Sample trajectories for the bottle and plate task. The policies with
        memory states succeed for each condition, and place the object close to
        the target location.
        \label{fig:cubby}
    }
    \vspace{-0.2in}
\end{figure}

All of our tasks involve some partially observed component, where the
observation $\ot$ received by the policy does not contain all of the information
necessary to accomplish the task, in contrast to the full state $\st$ that is
provided to the linear-Gaussian controllers during training. The simple
navigation task required the agent to travel to a designated position and, after
the first half of the episode, return to the starting location. The starting
location was varied, requiring a successful policy to remember its point of
origin in order to return there in the second half of the episode. This was
intentionally constructed to be sufficiently simple that the main challenge
stemmed from the partial observability, rather than the physical difficulty of
the behavior. The state space had two dimensions, and the two dimensional
actions directly set the agent's velocity in the plane.

In the more complex manipulation tasks, the policies needed to control a 7
degree of freedom robot arm directly with joint torques in a full physics
simulator. The controls $\at$ had 7 dimensions, and the configuration of the
robot was provided in terms of joint angles and two 3D points on the object
being manipulated, as well as their time derivatives, for a total dimensionality
of 26. In the first manipulation task, shown in Figure~\ref{fig:pegsort}, the
robot was required to sort a peg into one of two holes. The hole position was
provided to the policy on the first time step, and the policy was required to
remember this position and move to the correct target. To prevent the policy
from applying a large force in the direction of the target immediately when the
target was presented, the robot was not allowed to physically move until the
second time step. This task therefore could not be completed without memory, and
provides a good comparison between our method and the alternative architectures.

In the second manipulation task, shown in Figure~\ref{fig:cubby}, the robot was
required to insert plates and bottles through a horizontal or vertical slot
(``cubby''), and position them in the desired pose. The robot was required to
determine both which object it is holding, and which way the cubby is oriented
in order to angle the object correctly. This task is in fact possible to solve
without memory, by using an appropriate reactive strategy that responds to
collisions, but becomes significantly easier when memory is available.

\subsection{Results}

The results for each method on each of the tasks are presented in
Figure~\ref{fig:results}. Each of the graphs shows the distance to the target
for each task in terms of the number of samples used for training. For the
manipulation tasks, the distance is measured between the object and its desired
position, while for the navigation task, the reported distance is the larger of
the minimum distance to target in the first part of the episode and the minimum
distance to the initial state in the second part. For each task, we show
separate plots for each condition. For the peg sorting task, there are two
conditions corresponding to the two targets, and the dotted line shows the depth
of the hole. Policies with minimum distances above this line fail to insert the
peg into the hole. For the cubby task, the conditions correspond to the
orientation of the cubby and whether the robot is holding a plate or bottle. For
the navigation task, the conditions correspond to different starting states.
Each method was provided with 5 samples per target per iteration, and RWR was
also tested with 25 samples per iteration. Traces of the trajectories attained
by each method are shown under the corresponding plots.

Good performance on each task requires the policy to succeed for all of the
conditions. For the peg sorting task, our method is able to insert the peg into
the hole for both targets, while the feedforward policy simply picks the same
target each time, succeeding on one condition but failing on the other. The
standard LSTM also did not learn to remember the target, and instead found a
``middle ground'' strategy where it moved to the center rather than choosing a
hole. Despite the fact that in theory this network could complete this task, in
practice we found the LSTM network to be more difficult to train than our
method, which required substantially less tuning.\footnote{We tested a variety
of hyperparameters for the LSTM baseline and chose the best-performing policy.}
The hybrid network that consisted of feedforward and LSTM layers outperformed
both the feedforward and pure LSTM policy, but still did not achieve the same
performance as our memory states method.

On the bottle and plate task, the feedforward policy was able to succeed on
three of the four conditions, but was unable to rotate the bottle to insert it
into the horizontal cubby. Both the LSTM and hybrid policies were able to
successfully insert the object into the cubby, but the resulting policies were
substantially less stable, and were unable to position the object accurately at
the desired position. This again reflects the difficulty of optimizing recurrent
policies with backpropagation through time. In contrast, our policy with memory
states was able to both insert the object into the cubby in each condition, and
position it accurately at the target location. This task neatly illustrates one
of the motivating factors for our method: even without explicit memory states,
feedforward policies can adopt strategies that ``off\-load'' memory onto the
physical state of the system, by utilizing subtly different joint angles and
velocities depending on their past experience. However, with internal memory,
this type of physical ``offloading'' is unnecessary.

For both manipulation tasks, RWR was unable to discover an effective policy,
either with a neural network parameterization or with the linear
parameterization shown in the plots, though the linear variant achieved slightly
lower cost. This agrees with results reported in prior work \cite{la-lnnpg-14},
which showed that, for tasks of this type, guided policy search typically
outperformed direct policy search methods, including RWR.

For the 2D navigation and retrieval task, our method was able to succeed from
each of the starting positions. The feedforward network could not return the
object back to the initial state due to lack of memory, while both the standard
LSTM and hybrid policies could not be optimized successfully and did not produce
a coherent behavior. Due to the substantially lower dimensionality of this task,
RWR was in fact able to discover a policy that succeeded on one of the four
conditions, but could not learn to effectively utilize the memory states to
succeed from all four initial states.


The project website contains supplementary videos that illustrate the behavior
of these policies.\footnote{\url{ http://rll.berkeley.edu/gpsrnn/}}

\section{Discussion and Future Work}
\label{sec:discussion}

We presented a method for training policies for continuous control tasks that
require memory. We augment the state space of the system with memory states,
which the policy can choose to read from and write to as needed to accomplish
the task. In order to make it tractable for the policy to learn effective
memorization and recall strategies, we use guided policy search, which employs a
simple trajectory-centric reinforcement learning algorithm to optimize over the
memory state activations. This trajectory optimization procedure effectively
tells the policy which information needs to be stored in the memory states, and
the policy only needs to figure out how to reproduce the memory state
activations. This tremendously simplifies the problem of searching over
memorization strategies in comparison to model-free policy search methods and,
unlike standard model-based methods for recurrent policies, it also avoids the
need to backpropagate the gradient through time. However, when viewed together
with the memory states, the policy is endowed with memory, and can be regarded
as a recurrent neural network. Our experimental results show that our method can
be used to learn policies for a variety of simulated robotic tasks that require
maintaining internal memory to succeed.

Part of the motivation for our approach came from the observation that even
fully feedforward neural network policies could often complete tricky tasks that
seemed to require memory by using the physical state of the robot to ``store''
information, similarly to how a person might ``remember'' a number while
counting by using their fingers. In our approach, we exploit this capability of
reactive feedforward policies by providing extra state variables that do not
have a physical analog, and exist only for the sake of memory.

While we presented experiments in simulation, guided policy search has been
applied extensively on real robotic platforms \cite{lwa-lnnpg-15,levine2015end},
and the modifications proposed in this paper do not introduce additional
complications into real-world robotic applications. Experiments on a real-world
robotic platform would be valuable for evaluating the degree to which memory
states can help general-purpose policies deal with partial observability
stemming from real-world sensors, such as occlusions in camera images.

Another interesting direction for follow-up work is to apply our approach for
training recurrent networks for general supervised learning tasks, rather than
just robotic control. In this case, the memory state comprises the entire state
of the system, and the cost function is simply the supervised learning loss.
Since the hidden memory state activations are optimized separately from the
network weights, such an approach could in principle be more effective at
training networks that perform complex reasoning over temporally extended
intervals. Furthermore, since our method trains stochastic policies, it would
also be able to train stochastic recurrent neural networks, where the transition
dynamics are non-deterministic. These types of networks are typically quite
challenging to train, and exploring this further is an exciting direction for
future work.

\newpage
{\small
\bibliographystyle{ieeetran}
\bibliography{references}
}

\end{document}